\documentclass{bmvc2k}

\ifdefined\camerareadyclean\else
    \def\proofreadreview{1}
\fi

\usepackage{graphicx}
\usepackage{amsmath,amssymb}
\usepackage{booktabs}
\usepackage{comment}
\usepackage{multirow}
\usepackage[table]{xcolor}
\usepackage{pgfplots}
\usepgfplotslibrary{groupplots}
\pgfplotsset{compat=1.18}
\usepackage{url}

\definecolor{plotgray}{HTML}{B8BEC5}
\definecolor{rawlight}{HTML}{56B4E9}
\definecolor{rawdark}{HTML}{0072B2}
\definecolor{rgblight}{HTML}{F4B942}
\definecolor{rgbmid}{HTML}{E6862A}
\definecolor{rgbdark}{HTML}{C84D0A}
\definecolor{targetgreen}{HTML}{2CA25F}

\newcommand{\sta}{{\setlength{\fboxsep}{1pt}\colorbox{green!8}{\textbf{Strategy A}}}\xspace}

\newcommand{\stb}{{\setlength{\fboxsep}{1pt}\colorbox{red!8}{\textbf{Strategy B}}}\xspace}

\ifdefined\proofreadreview
    \usepackage[normalem]{ulem}
    
    \DeclareRobustCommand{\revadd}[1]{\textcolor{red}{#1}}
\else
    
    \providecommand{\revadd}[1]{#1}
\fi

\title{Benchmarking RAW and RGB Restoration in Image Signal Processors}

\addauthor{Zihao Lu}{zihao.lu@stud-mail.uni-wuerzburg.de}{1}
\addauthor{Radu Timofte}{radu.timofte@uni-wuerzburg.de}{1}
\addauthor{Marcos V. Conde~$^{\dagger}$}{marcos.conde@uni-wuerzburg.de}{1}

\addinstitution{
 Computer Vision Laboratory\\
 CAIDAS and IFI \\
 University of Würzburg\\
 Würzburg, Germany\\[1em]
 \small{$^{\dagger}$~Corresponding Author}\\
 \small{\texttt{https://github.com/mv-lab/AISP}}
}

\runninghead{Lu, Timofte and Conde}{Benchmarking RAW-RGB Restoration for ISP}

\hypersetup{
  pdftitle  = {Benchmarking RAW and RGB Restoration in Image Signal Processors},
  pdfauthor = {Zihao Lu, Radu Timofte, Marcos V. Conde},
  pdfsubject = {BMVC 2026},
  pdfkeywords = {benchmarking, Raw restoration, image signal processor}
}

\begin{document}

\maketitle

\begin{figure*}[ht]
    \centering
    \includegraphics[width=\linewidth]{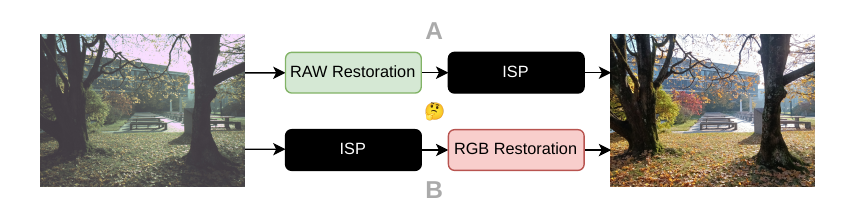}
    \caption{\textbf{Where should restoration happen in camera pipelines?} Considering the limited computational budget in mobile devices, we compare two strategies around a fixed Image Signal Processor: (A) pre-ISP RAW restoration and (B) post-ISP RGB restoration.
    }
    \label{fig:teaser}
\end{figure*}

\begin{abstract}

Modern cameras transform RAW sensor measurements into sRGB images through an image signal processor (ISP). We benchmark two placements for blind restoration around a fixed ISP: (A) pre-ISP restoration in the RAW domain and (B) post-ISP restoration in the sRGB domain. The benchmark covers four smartphone device groups, two learned ISPs, three degradation regimes---noise, blur, and joint noise and blur---, and several representative RAW and RGB restoration models.

Our results show that placement alone does not determine performance. The RAW restoration strategy outperforms the best generic RGB restoration models. However, RGB restoration models trained considering the ISP transformations, achieve the best overall performance. Our novel benchmark demonstrates that the image reconstruction performance strongly depends on the
alignment between the restoration model and the target imaging pipeline. We consequently recommend reporting restoration placement and ISP-aware supervision as key experimental factors.
\end{abstract}


\section{Introduction and Motivation}
\label{sec:intro}

Image restoration aims to recover high-quality images from degraded observations, such as noise, blur, and their joint effects. Most existing restoration methods operate in the RGB domain, where large-scale datasets and mature network architectures have led to strong performance on standard restoration benchmarks. However, RGB images are not direct sensor measurements. Digital cameras and smartphones employ a dedicated Image Signal Processor (ISP) to transform RAW sensor measurements into visually pleasing RGB images.

\begin{figure}
    \centering
    \includegraphics[width=\linewidth]{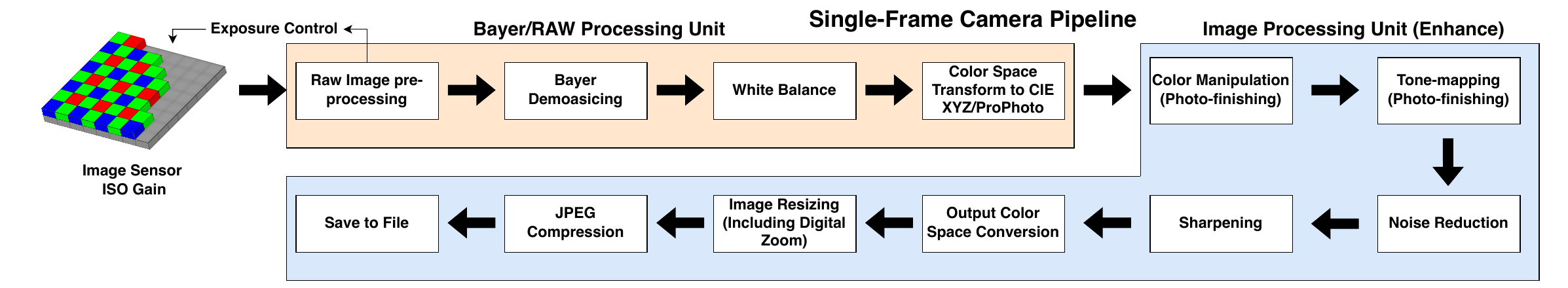}
    \caption{A classical Image Signal Processor~\cite{delbracio2021mobile}.}
    \label{fig:isp-teaser}
\end{figure}

A simplified ISP pipeline can be expressed mathematically as:
\begin{equation}
    \mathbf{y}_{rgb} = \text{ISP}(\mathbf{x}_{raw}) = \gamma(\mathbf{C}(\mathbf{W}(\mathcal{D}(\mathbf{x}_{raw}))))
\end{equation}
where $\mathbf{x}_{raw} \in \mathbb{R}^{H \times W}$ represents mosaiced RAW sensor data (typically following a Bayer RGGB pattern), and $\mathbf{y}_{rgb} \in \mathbb{R}^{H \times W \times 3}$ is the final sRGB image. The core operations include (i) demosaicing $\mathcal{D}(\cdot)$ to interpolate missing color values, (ii) white balancing $\mathbf{W}$ to correct color temperature, (iii) a color correction matrix (CCM) $\mathbf{C}$ for device-specific color calibration, and (iv) tone mapping $\gamma(\cdot)$ to compress dynamic range and apply gamma correction. Figure~\ref{fig:isp-teaser} illustrates this camera pipeline.

In practice, white balance and CCM parameters are typically calibrated or tuned based on sensor characterization and color charts~\cite{karaimer2016software}. The tone mapping $\gamma(\cdot)$ is highly nonlinear (often a power function with $\gamma \approx 1/2.2$ plus S-curves), which can amplify noise in darker regions and create complex interactions between degradations. Consequently, post-ISP restoration must handle artifacts that have been transformed by the camera pipeline. Conversely, a post-ISP model trained on outputs from the target pipeline can learn these ISP-specific artifacts directly.

These observations motivate a controlled comparison of restoration before and after the ISP. Existing comparisons often confound restoration domain with degradation synthesis, training data, camera pipeline, or evaluation protocol. It therefore remains unclear whether performance differences arise from the RAW or RGB domain itself or from alignment between the restoration training distribution and the target pipeline.

To compare RAW pre-processing and RGB post-processing, we assume that \emph{\textbf{the ISP is fixed and treated as a black box}}: its weights are never updated jointly with the restoration model, and the restoration model does not use ISP parameters or gradients. This reflects proprietary camera pipelines whose internal operations are inaccessible. Recent work has explored learning ISP transformations end-to-end~\cite{ignatov2020replacing, punnappurath2022day, conde2022model}, but practical challenges remain: (i) large-scale paired RAW--RGB training data are expensive and device-specific; (ii) real camera ISPs are commonly proprietary and non-differentiable; and (iii) their behavior varies with lighting, scene content, and exposure. A restoration model may nevertheless be trained on examples from the output distribution of a fixed target ISP without accessing its internal parameters and configuration. 

As shown in Figure~\ref{fig:teaser}, \sta always denotes pre-ISP RAW restoration, whereas \stb always denotes post-ISP RGB restoration. Independently of placement, we distinguish \emph{generic} models, which are not trained on outputs from the target ISP, from \emph{ISP-aware} models, which are trained using examples from the target fixed ISP's output distribution. Under this terminology, RAW restoration is stronger than generic RGB restoration (using all-in-one pre-trained models), while ISP-aware RGB restoration performs best in the sensor-specific comparison.

\vspace{-2mm}
\paragraph{Contributions} 
In this paper, we introduce a \textbf{systematic comparison framework} for evaluating \emph{pre-ISP} RAW restoration and \emph{post-ISP} RGB restoration with fixed ISPs. The framework separates restoration placement from the training regime, allowing us to distinguish domain effects from alignment with the target ISP distribution.

Our work fills this gap through a \textbf{multi-camera evaluation} spanning four smartphone groups and three degradation levels (denoising, deblurring, and joint restoration), together with real-world examples.

Our \textbf{experimental validation} shows that restoration performance depends strongly on the training distribution. Generic pre-trained RGB methods struggle with ISP-transformed degradations, whereas an RGB model trained using outputs from the \emph{fixed target ISP} can recover both input degradations and pipeline-specific artifacts. This finding qualifies any universal claim that either RAW or RGB restoration is intrinsically preferable.

\section{Related Work}
\label{sec:relate}

\subsection{Learned Image Signal Processing}

Traditional ISPs rely on carefully tuned pipelines with handcrafted parameters~\cite{karaimer2016software}. Recent work has explored replacing these with end-to-end learned models that directly map RAW sensor data to sRGB images~\cite{schwartz2018deepisp, zamir2020cycleisp, liang2021cameranet, xing2021invertible, conde2022model}. While these methods demonstrate impressive results under controlled conditions, they face practical limitations: training requires extensive paired RAW-RGB datasets that are sensor-specific~\cite{ignatov2020replacing, ignatov2022learned}, and\revadd{,} crucially\revadd{,} \emph{robustness to input degradations is rarely evaluated}. Our work instead treats the ISP as a fixed component and benchmarks modular restoration before and after it.

\subsection{Image Restoration}
\paragraph{RGB Image Restoration}

Deep learning has enabled remarkable progress in blind image restoration. Task-specific methods target denoising~\cite{zhang2017dncnn, zhang2022practical}, deblurring~\cite{nah2017Gopro, hosseini2019convolutional}, and super-resolution~\cite{zhang2021designing}, while all-in-one networks~\cite{li2022all, vaishnav2023promptir, kong2024towards, conde2024instructir} handle multiple types of degradation using learned prompts or contrastive learning.

When applied to camera pipelines, generic RGB restoration methods face a distribution mismatch. They operate on nonlinear sRGB images after tone mapping and other device-specific processing, which changes the appearance of noise, blur, and content interactions~\cite{brooks2019unprocessing, delbracio2021mobile}. A model trained on conventional RGB degradations may therefore generalize poorly to outputs from a particular ISP. This motivates both restoration \emph{before} the ISP, where degradations are better characterized, and target-ISP training for restoration \emph{after} the ISP.

\paragraph{RAW Image Restoration}

Most prior works focus on RAW denoising~\cite{hasinoff2016burst, abdelhamed2018high, brooks2019unprocessing, wang2020practical} or deblurring in controlled settings~\cite{liang2020raw}, while RawSR~\cite{xu2019rawsr} and BSRAW~\cite{conde2024bsraw} tackle RAW super-resolution with degradation models. Although these works demonstrate benefits of RAW-domain processing, they do not systematically compare RAW pre-processing and RGB post-processing for smartphone ISPs under different training regimes. Existing RAW restoration work also commonly assumes knowledge of the downstream ISP or jointly trains the restoration and ISP. We instead compare modular restoration around fixed ISPs and analyze how the target training distribution changes the observed RAW--RGB ranking.

\section{Methodology}
\label{sec:method}

\subsection{Dataset}
\label{sec:dataset}

Our goal is blind restoration of RAW images captured by smartphone cameras. Because obtaining pixel-aligned degraded--clean RAW pairs in real scenes is difficult, we synthesize training data by applying controlled noise and blur to clean RAW captures.

We use the RawIR dataset~\cite{conde2024toward, conde2025ntire}, comprising Vivo X90 Pro, iPhone XS, Samsung S9/S21, and Google Pixel 7--9 images. The fixed train/test counts are 452/10, 955/17, 465/9, and 220/11, respectively, totaling 2,092 training and 47 test images. The test images were sampled approximately in proportion to each device subset and manually checked for image quality and scene and content diversity. All methods and degradation levels use the same split, preventing test-set differences from affecting the comparisons. The relatively small test set remains a limitation when interpreting broad cross-sensor generalization.

The RAW pre-processing pipeline consists of three steps: (i) we normalize each RAW image using its sensor-specific black level and bit depth (typically 10--14 bits); (ii) we pack the mosaiced data into a four-channel RGGB representation, preserving the color filter array (CFA) structure; and (iii) following established RAW-processing practice~\cite{brooks2019unprocessing, xu2019rawsr}, we crop non-overlapping $512 \times 512 \times 4$ RAW patches, corresponding to $1024 \times 1024 \times 3$ RGB regions, for training.

\paragraph{Three-Level Degradation Protocol.}
Our benchmark includes test data at three degradation levels, each evaluating a specific restoration capability (see Table~\ref{tab:isp_on_deg_data}):

\begin{equation}
    \mathbf{y} = (\mathbf{x} \otimes \mathbf{k}) + \mathbf{n}
    \label{eq:deg_pipeline}
\end{equation}

\begin{itemize}
    \item \textbf{Level 1, Denoising:} Noise only: $\mathbf{y} = \mathbf{x} + \mathbf{n}$, where $\mathbf{n}$ is heteroscedastic Gaussian or real sensor noise modeled as $n \sim \mathcal{N}(0,\alpha x+\sigma_r^2)$ as previous works~\cite{hasinoff2016burst,brooks2019unprocessing,zamir2020cycleisp}
    
    \item \textbf{Level 2, Deblurring:} Blur only: $\mathbf{y} = \mathbf{x} \otimes \mathbf{k}$, where $\mathbf{k}$ represents a point-spread function (PSF) that models different types of defocus or motion blur~\cite{nah2017Gopro, nah2021ntire, hosseini2019convolutional}.
    
    \item \textbf{Level 3, Joint:} Blur and noise: $\mathbf{y} = (\mathbf{x} \otimes \mathbf{k}) + \mathbf{n}$, representing the most challenging real-world scenario~\cite{elad1997restoration,xu2019rawsr}.
\end{itemize}

During training, we randomize the degradation parameters, apply blur with probability 0.5, and sample the noise types with equal probability. We follow the degradation pipeline of prior works such as RawIR~\cite{conde2024toward, conde2025ntire}, using diverse noise profiles, PSFs, and blur kernels.

\begin{figure*}[t]
    \centering
    \setlength\tabcolsep{2pt}
    \resizebox{\linewidth}{!}{
        \begin{tabular} {c}
            \begin{tabular}{c c c}
                \includegraphics[width=0.32\linewidth]{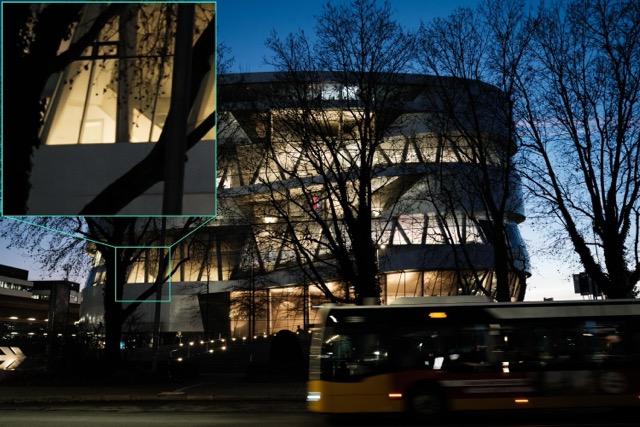} &
                \includegraphics[width=0.32\linewidth]{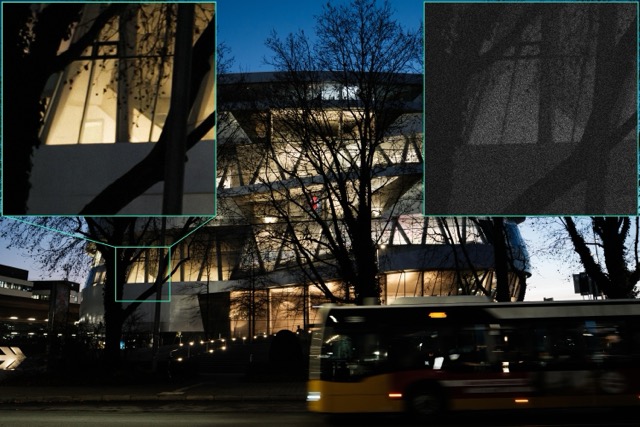} & 
                \includegraphics[width=0.32\linewidth]{BMVC2026/deg_pipeline/DSC06547/natural_noise_vis.jpeg } \\
                (a) Original Sample 1 & (b.1) Shot \& read noise & (b.2) Gaussian noise
            \end{tabular}
            \\
            \begin{tabular}{c c c}
                 \includegraphics[width=0.32\linewidth]{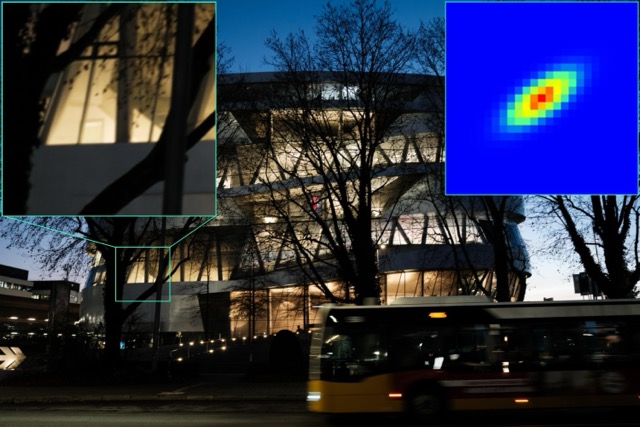} & 
                 \includegraphics[width=0.32\linewidth]{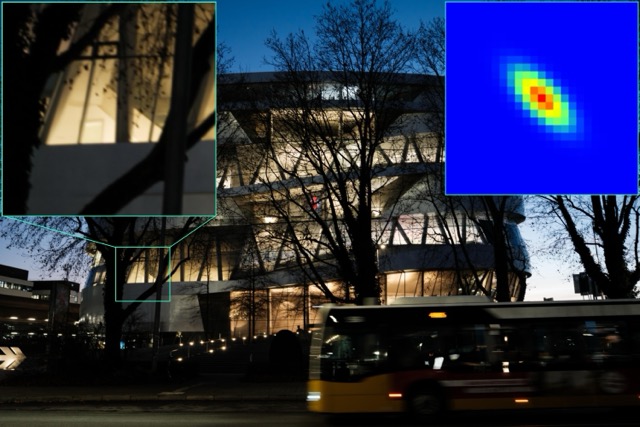} & 
                 \includegraphics[width=0.32\linewidth]{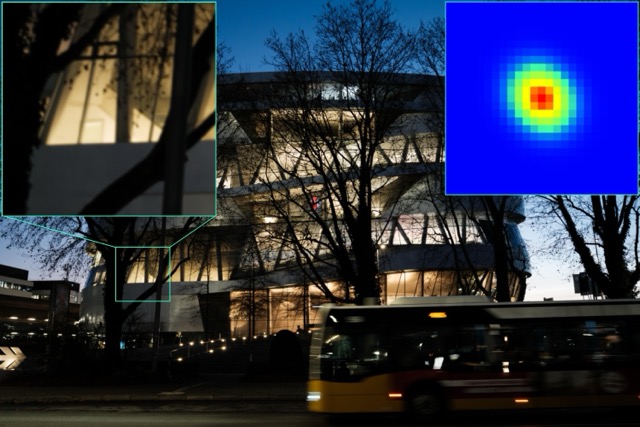} \\
                 (c.1) Linear Motion Blur & (c.2) Linear Motion Blur & (c.3) Linear Motion Blur \\
                 \includegraphics[width=0.32\linewidth]{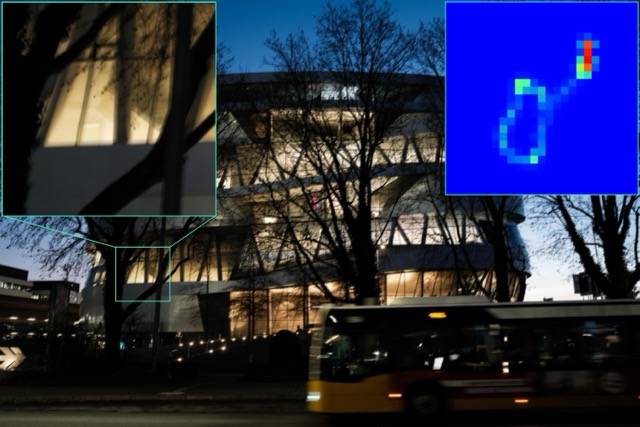} & 
                 \includegraphics[width=0.32\linewidth]{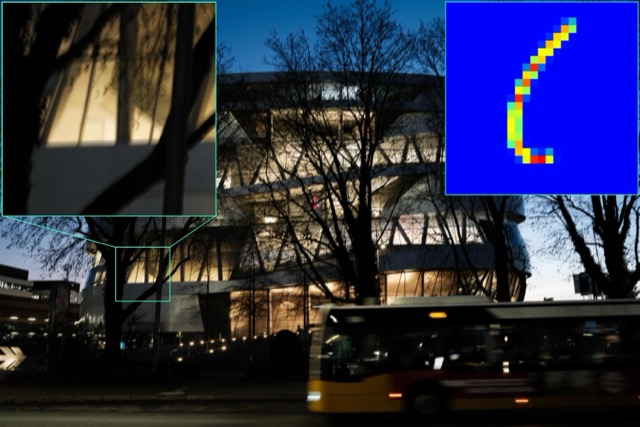} & 
                 \includegraphics[width=0.32\linewidth]{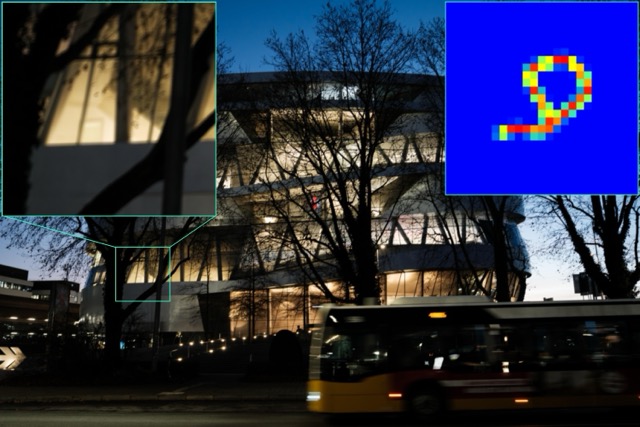} \\
                 (d.1) Complex Motion Blur & (d.2) Complex Motion Blur & (d.3) Complex Motion Blur \\
                 \includegraphics[width=0.32\linewidth]{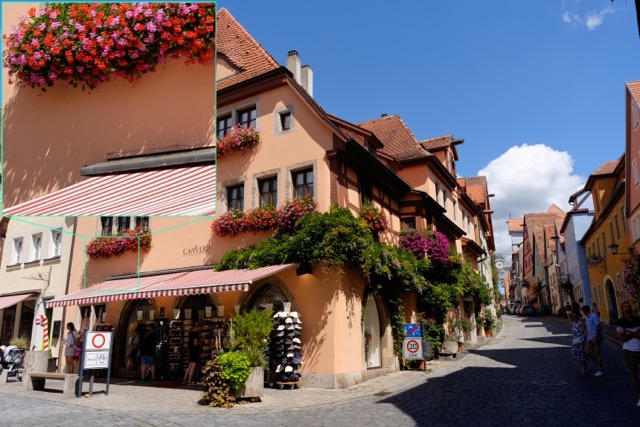} & 
                 \includegraphics[width=0.32\linewidth]{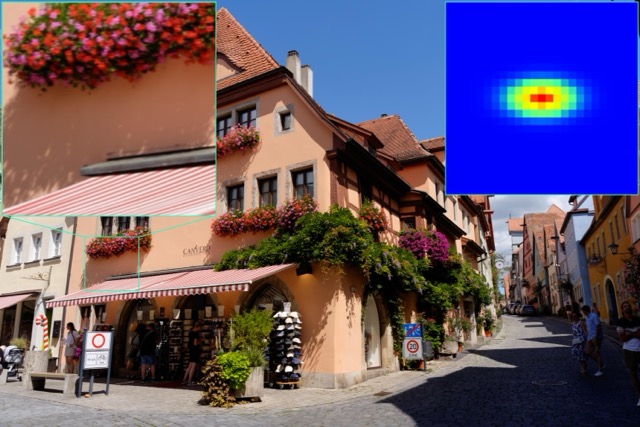} & 
                 \includegraphics[width=0.32\linewidth]{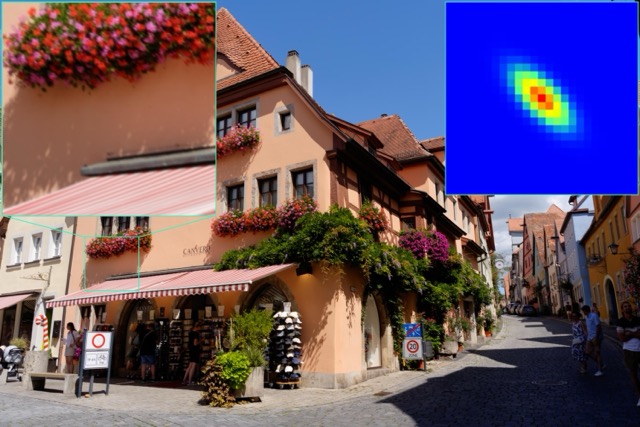} \\
                 (e) Original Sample 2 & (f.1) Linear Motion Blur & (f.2) Linear Motion Blur \\                 
                 \includegraphics[width=0.32\linewidth]{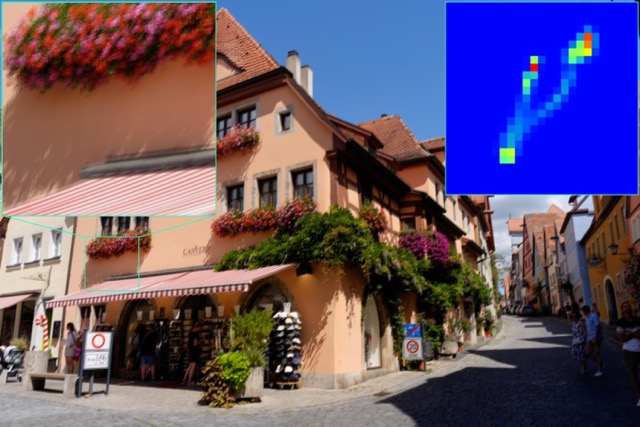} & 
                 \includegraphics[width=0.32\linewidth]{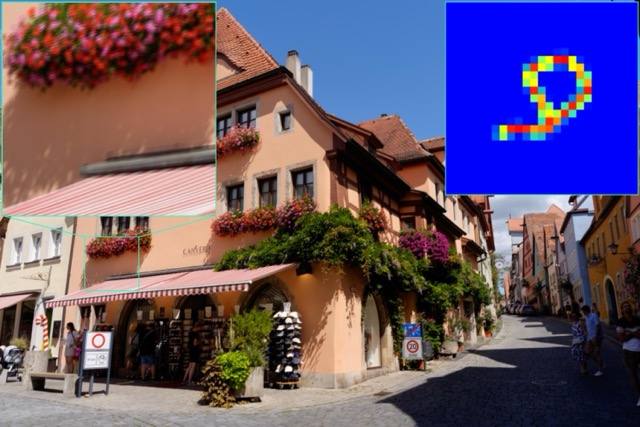} & 
                 \includegraphics[width=0.32\linewidth]{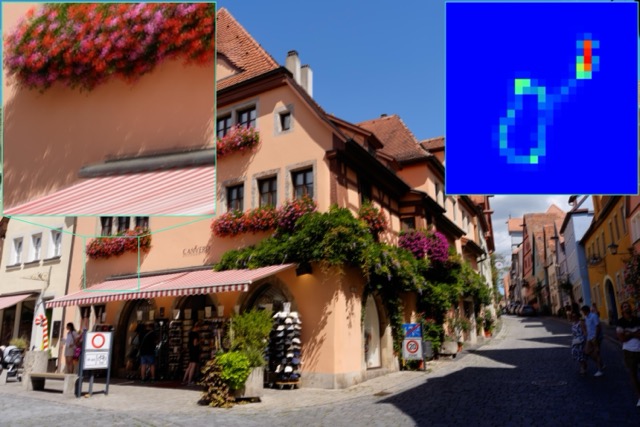} \\
                 (g.1) Complex Motion Blur & (g.2) Complex Motion Blur & (g.3) Complex Motion Blur \\
            \end{tabular}
        \end{tabular}
    }
    \caption{\textbf{Examples of synthetic degradations.} We apply diverse blur kernels and PSFs, visualized in the upper-right corner of each image, to approximate common blur patterns. Groups (a) and (b) illustrate a clean image and its noisy variants. Groups (c) and (f) show anisotropic Gaussian-like blur, whereas groups (d) and (g) show more complex motion blur. The RAW images are visualized after ISP processing.
    }
    \label{fig:deg_pipeline}
\end{figure*}


\begin{figure*}[t]
    \centering
    \setlength\tabcolsep{2pt}
    \resizebox{\linewidth}{!}{
        \begin{tabular}{c c}
            \includegraphics[width=0.5\linewidth]{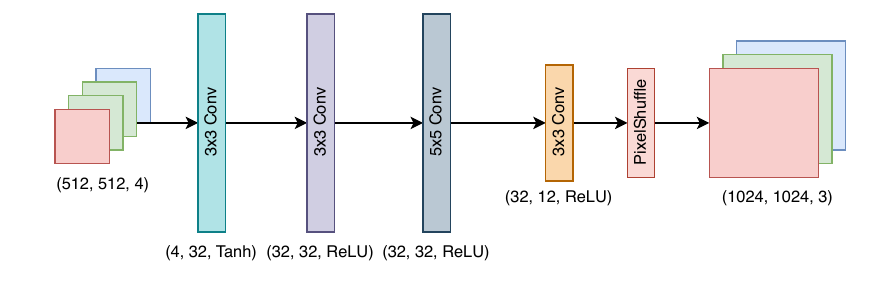} &
            \includegraphics[width=0.5\linewidth]{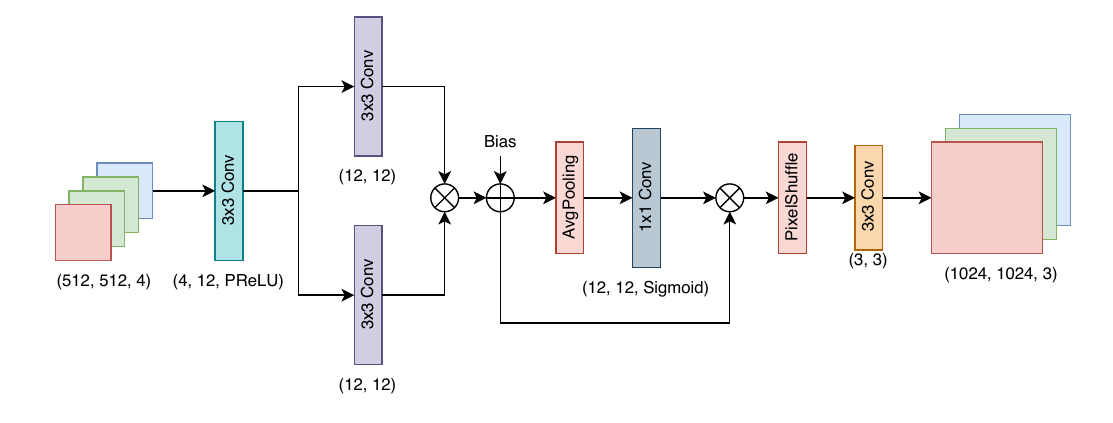} \\
            \textbf{ISP~v1} & \textbf{ISP~v2} \\
        \end{tabular}
    }
    \caption{\textbf{Two neural ISP architectures used as fixed pipeline proxies.} Left: a winning architecture from the Learned Smartphone ISP Challenge~\cite{ignatov2022learned}. Right: a winning architecture from the Mobile AI 2025 challenge~\cite{ignatov2025learned}.}
    \label{fig:isp}
\end{figure*}


\begin{table*}[!ht]
    \centering
    \resizebox{0.95\linewidth}{!}{
        \begin{tabular}{c | c | c | c | c | c}
            \hline
            \rowcolor{gray!15} Deg. Level &  ISP  &  Vivo X90    & Google Pixel  & Samsung S9  &   iPhone XS   \\
            \hline
            \rowcolor{yellow!5} \cellcolor{white} & ISP v1 & \textbf{22.60} / \textbf{0.90} & \textbf{22.31} / \textbf{0.88} & \textbf{29.67} / \textbf{0.96} & 17.56 / \textbf{0.81} \\
            \rowcolor{blue!5}   \cellcolor{white}\multirow{-2}{*}{\bf Clean} & ISP v2 & 18.97 / 0.81 & 19.75 / 0.80 & 27.95 / 0.94 & \textbf{17.58} / 0.77 \\
            \hline
            \rowcolor{yellow!5} \cellcolor{white} & ISP v1 & \textbf{21.52} / 0.77 & \textbf{21.04} / 0.75 & \textbf{26.31} / 0.83 & 17.08 / 0.66 \\
            \rowcolor{blue!5} \cellcolor{white}\multirow{-2}{*}{\bf Level 1} & ISP v2 & 21.26 / \textbf{0.82} & 19.52 / \textbf{0.77} & 25.72 / \textbf{0.84} & \textbf{17.45} / \textbf{0.73} \\
            \hline
            \rowcolor{yellow!5} \cellcolor{white} & ISP v1 & \textbf{20.90} / \textbf{0.79} & \textbf{19.17} / \textbf{0.67} & \textbf{22.48} / \textbf{0.81} & 16.01 / 0.67 \\
            \rowcolor{blue!5} \cellcolor{white}\multirow{-2}{*}{\bf Level 2} & ISP v2 & 20.10 / \textbf{0.79} & 17.83 / 0.64 & 22.08 / \textbf{0.81} & \textbf{17.11} / \textbf{0.70} \\
            \hline
            \rowcolor{yellow!5} \cellcolor{white} & ISP v1 & 19.37 / 0.63 & \textbf{17.67} / 0.49 & \textbf{22.31} / 0.73 & 15.85 / 0.55 \\
            \rowcolor{blue!5} \cellcolor{white}\multirow{-2}{*}{\bf Level 3} & ISP v2 & \textbf{19.43} / \textbf{0.70} & 17.58 / \textbf{0.57} & 21.86 / \textbf{0.74} & \textbf{16.94} / \textbf{0.66} \\
            \hline
        \end{tabular}
    }
    \caption{\textbf{Evaluation of neural ISPs} under different degradation levels. No restoration model is used. We report PSNR/SSIM metrics in the RGB domain. Best results are shown in bold; the same convention is used in the following tables.
    }
    \label{tab:isp_on_deg_data}
\end{table*}

\begin{table}[t]
    \centering
    \begin{tabular}{l | c | c | c | c | c}
    \hline
         \rowcolor{gray!15} & UNet~\cite{ronneberger2015u}  & NAFNet~\cite{chen2022simple} & PMRID~\cite{wang2020practical} & MOFA~\cite{chen2023mofa}  & MFDNet~\cite{jiang2024mfdnet} \\
    \hline
    PSNR & 37.44 & \textbf{39.70} & 38.43 & 38.71 & 37.10  \\
    SSIM & 0.962 & \textbf{0.972} & 0.965 & 0.966 & 0.957  \\
    \hline
    Params. (M) & \textbf{0.266}  & 1.130 & 1.032 & 0.971  & 2.179 \\
    MACs (G)   & 2.23    & 3.99  & 1.21  & \textbf{1.14}  & 6.88  \\
    \hline
    \end{tabular}
    \caption{Benchmark of \emph{pre-ISP} RAW restoration (\textbf{RAWRes}) models.}
    \label{tab:model_baseline}
\end{table}


\subsection{Neural ISP Models}


The goal of this work is to improve the robustness of an Image Signal Processor using a restoration network. Designing and training neural ISPs is a separate research problem~\cite{ignatov2022learned, ignatov2025learned}; here, the restoration models are treated as modular blocks placed before or after the ISP.

Because real smartphone ISPs are proprietary, we use learned ISP models as controlled proxies for generating and evaluating RGB outputs. Once trained, these proxy ISPs remain fixed in every restoration experiment.

We train two neural ISP architectures from the MAI Learned ISP Challenges~\cite{ignatov2022learned, ignatov2025learned} to render sRGB images from RAW inputs. Figure~\ref{fig:isp} illustrates both architectures. Each ISP is trained for a specific camera sensor using curated, aligned RAW--sRGB pairs, where the sRGB targets are produced by the corresponding phone pipeline.

Table~\ref{tab:isp_on_deg_data} shows that the neural ISPs are sensitive to degraded RAW inputs. On Vivo X90, for example, ISP~v1 reaches 22.60 dB on clean inputs. Its performance falls to 21.52 dB with noise, 20.90 dB with blur, and 19.37 dB with joint noise and blur, corresponding to drops of 1.08, 1.70, and 3.23 dB, respectively.


\section{Experimental Results}
\label{sec:exp}

Throughout our experiments, the neural ISP is fixed: no restoration method updates it or receives gradients through it. Strategy A restores the RAW input before this ISP, and Strategy B restores its RGB output. We use ISP-aware (``target-ISP-trained'') only when the restoration training pairs contain outputs from the same fixed ISP used at evaluation time.

\subsection{RAW Restoration Prior}

To benchmark \sta, RAW-domain restoration, we consider architectures ranging from foundational baselines to recent efficient restoration models~\cite{ronneberger2015u, chen2022simple, wang2020practical, chen2023mofa, liu2023lightweight, jiang2024mfdnet}. We include UNet~\cite{ronneberger2015u}, which is widely used in image restoration~\cite{conde2023perceptual, zhang2021plug, brooks2019unprocessing}, and PMRID~\cite{wang2020practical}, which targets sensor-specific RAW denoising.

We also train MFDNet~\cite{jiang2024mfdnet}, NAFNet~\cite{chen2022simple}, and MOFA~\cite{chen2023mofa}. To keep NAFNet comparable in size to the other baselines, we use a reduced version with three encoder blocks, one middle block, three decoder blocks, and simplified skip connections. 
All models are trained on degraded--clean RAW pairs generated by the controlled degradation pipeline and organized by camera sensor. They receive normalized four-channel RGGB inputs corrected for black and white levels and produce restored RAW images, which are then passed to the fixed ISPs.

Table~\ref{tab:model_baseline} demonstrates that NAFNet and MOFA outperform other methods while maintaining reasonable parameter counts and computational complexity, so we use these two methods in the subsequent evaluation.

\begin{table*}[!ht]
    \centering
    \resizebox{0.95\linewidth}{!}{
        \includegraphics[]{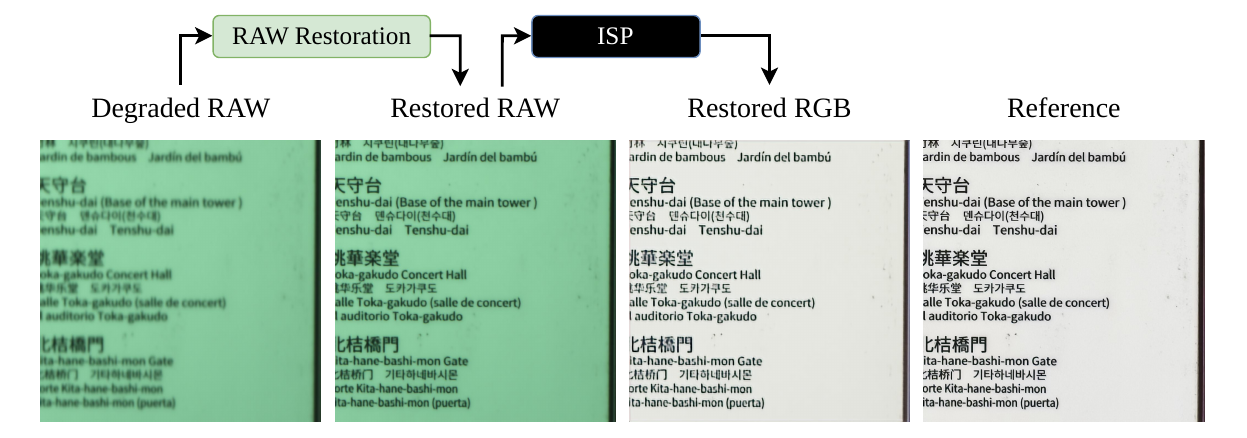}
    }
    \resizebox{0.95\linewidth}{!}{
        \begin{tabular}{c | c | c | c | c | c | c}
            \hline
            \rowcolor{gray!15} Deg. Level & Res. Model &  ISP  &  Vivo X90    & Google Pixel  & Samsung S9  &   iPhone XS   \\
            \hline
            \rowcolor{yellow!5} \cellcolor{white}                         & MOFA              & ISP v1 & \textbf{22.06} / \textbf{0.85} & 21.40 / 0.80 & 27.70 / 0.91 & 16.11 / 0.71 \\
            \rowcolor{yellow!5} \cellcolor{white}                         & NAFNet            & ISP v1 & 22.00 / 0.85 & \textbf{21.48} / \textbf{0.81} & \textbf{27.93} / \textbf{0.92} & 16.16 / 0.72 \\
            \rowcolor{blue!5}   \cellcolor{white}                         & MOFA              & ISP v2 & 21.39 / 0.85 & 19.43 / 0.76 & 26.57 / 0.90 & \textbf{17.42} / \textbf{0.74} \\
            \rowcolor{blue!5}   \cellcolor{white}\multirow{-4}{*}{\bf Level 1} & NAFNet            & ISP v2 & 21.35 / 0.85 & 19.48 / 0.77 & 26.80 / 0.91 & 17.30 / 0.74 \\
            \hline
            \rowcolor{yellow!5} \cellcolor{white}                         & MOFA              & ISP v1 & 21.54 / 0.81 & 19.71 / 0.69 & 24.58 / 0.85 & 15.75 / 0.67 \\
            \rowcolor{yellow!5} \cellcolor{white}                         & NAFNet            & ISP v1 & \textbf{21.93} / \textbf{0.83} & \textbf{20.21} / \textbf{0.72} & \textbf{25.99} / \textbf{0.88} & 15.78 / 0.68 \\
            \rowcolor{blue!5}   \cellcolor{white}                         & MOFA              & ISP v2 & 20.77 / 0.81 & 18.42 / 0.67 & 24.03 / 0.84 & 17.14 / 0.71 \\
            \rowcolor{blue!5}   \cellcolor{white}\multirow{-4}{*}{\bf Level 2} & NAFNet            & ISP v2 & 21.12 / 0.83 & 18.81 / 0.70 & 25.25 / 0.87 & \textbf{17.18} / \textbf{0.71} \\
            \hline
            \rowcolor{yellow!5} \cellcolor{white}                         & MOFA              & ISP v1 & 20.76 / 0.78 & 19.12 / 0.65 & 24.43 / 0.83 & 15.60 / 0.65 \\
            \rowcolor{yellow!5} \cellcolor{white}                         & NAFNet            & ISP v1 & \textbf{21.10} / \textbf{0.80} & \textbf{19.77} / \textbf{0.69} & \textbf{25.41} / \textbf{0.85} & 15.66 / 0.66 \\
            \rowcolor{blue!5}   \cellcolor{white}                         & MOFA              & ISP v2 & 20.32 / 0.79 & 18.26 / 0.64 & 23.76 / 0.83 & 17.00 / 0.69 \\
            \rowcolor{blue!5}   \cellcolor{white}\multirow{-4}{*}{\bf Level 3} & NAFNet            & ISP v2 & 20.55 / 0.80 & 18.57 / 0.67 & 24.58 / 0.85 & \textbf{17.01} / \textbf{0.70} \\
            \hline
        \end{tabular}
    }
    \caption{\textbf{Benchmark of Strategy A:} RAW restoration followed by a fixed ISP. The restoration models are trained on degraded--clean RAW pairs without using downstream ISP outputs during training. Compared with Table~\ref{tab:isp_on_deg_data}, they improve robustness across the evaluated degradation levels.
    }
    \label{tab:robisp}
\end{table*}

\begin{table*}[!ht]
    \centering
    \resizebox{\linewidth}{!}{
        \includegraphics[]{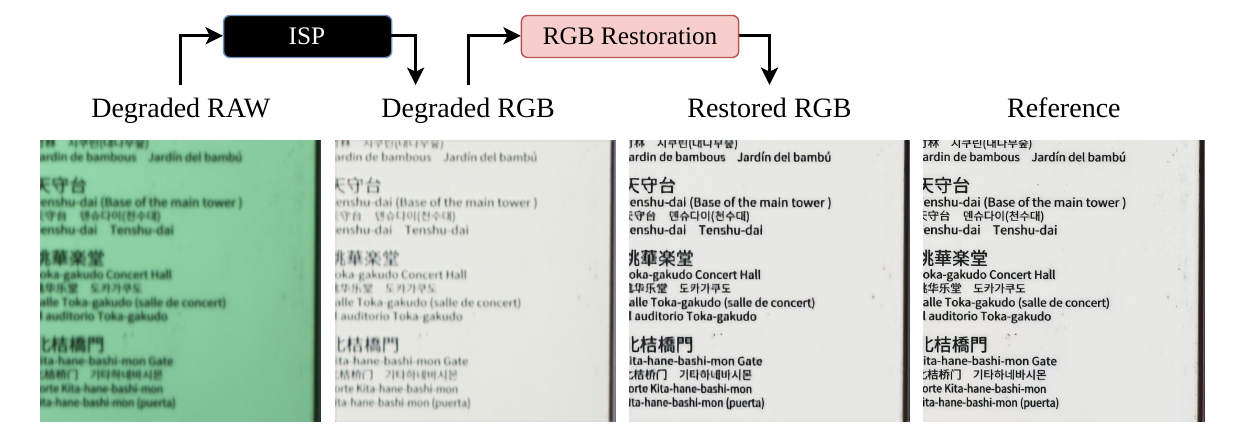}
    }
    \resizebox{0.95\linewidth}{!}{
        \begin{tabular}{c | c | c| c| c| c| c}
            \hline
            \rowcolor{gray!15} Deg. Level & Res. Model &  ISP  &  Vivo X90    & Google Pixel  & Samsung S9  &   iPhone XS   \\
            \hline
            \rowcolor{yellow!5} \cellcolor{white}                         & PromptIR          & ISP v1 & 20.46 / 0.75 & 20.58 / 0.74 & 25.36 / 0.82 & 16.63 / 0.64 \\
            \rowcolor{yellow!5} \cellcolor{white}                         & AirNet            & ISP v1 & 20.87 / 0.76 & \textbf{20.90} / 0.75 & 25.65 / 0.82 & 16.55 / 0.64 \\
            \rowcolor{yellow!5} \cellcolor{white}                         & MiOIR             & ISP v1 & \textbf{21.55} / 0.78 & 20.09 / 0.67 & \textbf{26.15} / 0.84 & \textbf{17.14} / 0.67 \\
            \rowcolor{blue!5}   \cellcolor{white}                         & PromptIR          & ISP v2 & 20.30 / 0.80 & 19.47 / \textbf{0.77} & 23.77 / 0.80 & 15.16 / 0.68 \\
            \rowcolor{blue!5}   \cellcolor{white}                         & AirNet            & ISP v2 & 20.52 / \textbf{0.81} & 19.32 / \textbf{0.77} & 25.03 / 0.83 & 16.22 / \textbf{0.70} \\
            \rowcolor{blue!5}   \cellcolor{white}\multirow{-6}{*}{\bf Level 1} & MiOIR             & ISP v2 & 20.57 / 0.72 & 19.38 / 0.72 & 25.92 / \textbf{0.85} & 16.52 / 0.57 \\
            \hline
            \rowcolor{yellow!5} \cellcolor{white}                         & PromptIR          & ISP v1 & 19.85 / 0.77 & 18.81 / \textbf{0.67} & 20.71 / 0.75 & 14.28 / 0.62 \\
            \rowcolor{yellow!5} \cellcolor{white}                         & AirNet            & ISP v1 & 19.25 / 0.75 & 18.09 / 0.64 & 19.95 / 0.71 & 12.87 / 0.59 \\
            \rowcolor{yellow!5} \cellcolor{white}                         & MiOIR             & ISP v1 & \textbf{20.86} / \textbf{0.79} & \textbf{19.22} / \textbf{0.67} & \textbf{22.49} / \textbf{0.81} & 16.11 / 0.67 \\
            \rowcolor{blue!5}   \cellcolor{white}                         & PromptIR          & ISP v2 & 19.62 / 0.78 & 17.75 / 0.64 & 21.34 / 0.79 & 13.98 / 0.64 \\
            \rowcolor{blue!5}   \cellcolor{white}                         & AirNet            & ISP v2 & 17.87 / 0.73 & 17.05 / 0.62 & 19.62 / 0.73 & 14.06 / 0.64 \\
            \rowcolor{blue!5}   \cellcolor{white}\multirow{-6}{*}{\bf Level 2} & MiOIR             & ISP v2 & 20.07 / \textbf{0.79} & 17.24 / 0.63 & 22.10 / \textbf{0.81} & \textbf{17.05} / \textbf{0.70} \\
            \hline
            \rowcolor{yellow!5} \cellcolor{white}                         & PromptIR          & ISP v1 & 18.70 / 0.61 & 17.36 / 0.48 & 21.51 / 0.71 & 15.60 / 0.53 \\
            \rowcolor{yellow!5} \cellcolor{white}                         & AirNet            & ISP v1 & 18.81 / 0.62 & 17.48 / 0.48 & 21.45 / 0.72 & 15.36 / 0.53 \\
            \rowcolor{yellow!5} \cellcolor{white}                         & MiOIR             & ISP v1 & \textbf{19.41} / 0.65 & \textbf{17.58} / 0.47 & 21.83 / 0.74 & 15.35 / 0.55 \\
            \rowcolor{blue!5}   \cellcolor{white}                         & PromptIR          & ISP v2 & 18.58 / 0.68 & 17.49 / \textbf{0.57} & 20.82 / 0.70 & 14.90 / \textbf{0.63} \\
            \rowcolor{blue!5}   \cellcolor{white}                         & AirNet            & ISP v2 & 18.81 / \textbf{0.69} & 17.34 / 0.56 & 21.13 / 0.73 & 15.61 / \textbf{0.63} \\
            \rowcolor{blue!5}   \cellcolor{white}\multirow{-6}{*}{\bf Level 3} & MiOIR             & ISP v2 & 18.97 / 0.61 & 17.35 / 0.52 & \textbf{21.94} / \textbf{0.75} & \textbf{16.45} / 0.57 \\
            \hline
        \end{tabular}
    }
    \caption{\textbf{Benchmark of generic Strategy B:} a fixed ISP followed by RGB restoration models pre-trained on conventional RGB degradations. Some models improve the degraded ISP outputs in Table~\ref{tab:isp_on_deg_data}, but they remain below the benchmark-trained Strategy A models in Table~\ref{tab:robisp}. This comparison is not training-matched.}
    \label{tab:robrgbisp}
\end{table*}

\subsection{RGB Restoration: Pre-trained Models}
\label{sec:rgb_res}

To benchmark generic \stb (\emph{post-ISP} restoration), we employ state-of-the-art \emph{all-in-one RGB restoration models} that can handle multiple types of degradation simultaneously.

We select three representative methods: AirNet~\cite{li2022all}; PromptIR~\cite{vaishnav2023promptir}, which is an all-in-one model based on Restormer~\cite{zamir2022restormer}; and MiOIR~\cite{kong2024towards}. These models receive degraded 8-bit sRGB images produced by the fixed ISP and return restored sRGB outputs. At this stage, ISP operations have transformed the original RAW-domain degradation characteristics.

These RGB models were pre-trained on large-scale RGB datasets with conventional synthetic degradations, rather than on outputs from our target ISPs. Despite their larger model sizes (15--30M parameters), they underperform the RAW models trained on our benchmark degradations across the evaluated levels (Table~\ref{tab:robrgbisp}). This is a practical comparison of generic pre-trained RGB models against benchmark-trained RAW models, rather than a training-matched isolation of the restoration domain.

These results indicate a gap between ISP-transformed degradations and the conventional RGB degradations used to pre-train these models. We therefore next evaluate training on a specific target pipeline.

\subsection{Sensor-Specific Models}
\begin{figure*}[!ht]
    \centering
    \begin{tikzpicture}
        \begin{axis}[
            width=0.88\linewidth,
            height=6.0cm,
            ymin=20.3,
            ymax=28.35,
            symbolic x coords={L1,L2,L3},
            xtick=data,
            xticklabels={L1 Noise,L2 Blur,L3 Joint},
            xticklabel style={font=\scriptsize},
            tick label style={font=\scriptsize},
            ylabel={PSNR (dB)},
            ylabel style={font=\small},
            axis line style={black!70},
            tick style={black!70},
            ymajorgrids,
            grid style={black!12, dashed},
            enlarge x limits=0.12,
            line width=1pt,
            mark size=2.2pt,
            clip=false,
            legend style={
                draw=none,
                fill=none,
                font=\small,
                legend columns=2,
                row sep=1pt,
                column sep=12pt,
                legend cell align={left},
                at={(0.5,-0.20)},
                anchor=north
            }
        ]
            \addplot[rawlight, densely dotted, line width=1.2pt, mark=o, mark options={fill=white, line width=0.9pt}] coordinates {(L1,22.06) (L2,21.54) (L3,20.76)};
            \addplot[rawdark!70, dashdotted, line width=1.2pt, mark=square*, mark options={fill=rawdark!70}] coordinates {(L1,22.00) (L2,21.93) (L3,21.10)};
            \addplot[
                rawdark,
                line width=1.5pt,
                mark=diamond*,
                nodes near coords={\pgfmathprintnumber[fixed,precision=2,zerofill]{\pgfplotspointmeta}},
                every node near coord/.append style={font=\scriptsize\bfseries, text=rawdark, anchor=south, yshift=2pt}
            ] coordinates {(L1,22.16) (L2,21.99) (L3,21.10)};
            \addplot[
                targetgreen,
                very thick,
                mark=triangle*,
                mark size=2.8pt,
                nodes near coords={\pgfmathprintnumber[fixed,precision=2,zerofill]{\pgfplotspointmeta}},
                every node near coord/.append style={font=\scriptsize\bfseries, text=targetgreen!70!black, anchor=south, yshift=2pt}
            ] coordinates {(L1,27.54) (L2,25.63) (L3,24.43)};
            \legend{Strategy A: MOFA (multi-sensor),Strategy A: NAFNet (multi-sensor),Strategy A: NAFNet (X90 RAW),Strategy B: NAFNet (X90 RGB; ISP-aware)}
        \end{axis}
    \end{tikzpicture}
    \caption{\textbf{Target-distribution comparison on Vivo X90 using fixed ISP~v1.} Blue curves are Strategy A RAW models evaluated after the ISP; the green curve is Strategy B trained directly on RGB outputs from that ISP. Sensor-specific RAW training changes performance only slightly, whereas ISP-aware RGB restoration yields the largest gain.}
    \label{fig:x90compare}
\end{figure*}
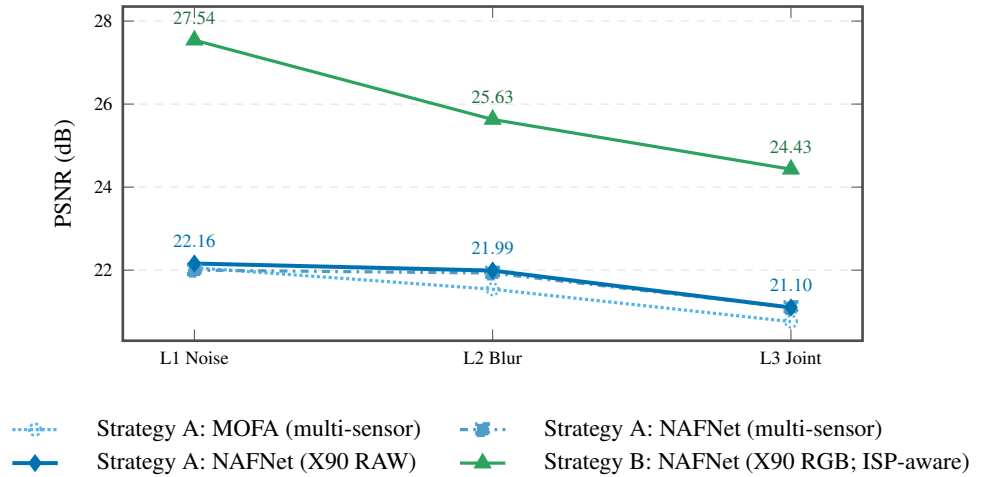

We additionally train comparable NAFNet~\cite{chen2022simple} models on the Vivo X90 data. The sensor-specific Strategy A model learns degraded-to-clean RAW restoration and is then followed by the fixed ISP. The ISP-aware Strategy B model learns from RGB pairs drawn from the output distribution of that same fixed ISP; it does not access ISP parameters or gradients. Figure~\ref{fig:x90compare} shows that the latter achieves higher RGB PSNR. Because the two losses operate in different domains, this experiment should not be interpreted as isolating domain alone. Instead, it demonstrates the benefit of training directly on the target output distribution.

\begin{figure*}[!ht]
    \centering
    \begin{tikzpicture}
        \begin{axis}[
            width=0.97\linewidth,
            height=6.0cm,
            ybar,
            /pgf/bar width=11pt,
            ymin=17,
            ymax=22.35,
            xmin=-3,
            xmax=15,
            xtick={0,6,12},
            xticklabels={L1 Noise,L2 Blur,L3 Joint},
            xticklabel style={font=\scriptsize},
            tick label style={font=\scriptsize},
            ylabel={PSNR (dB)},
            ylabel style={font=\small},
            axis line style={black!70},
            tick style={black!70},
            ymajorgrids,
            grid style={black!12, dashed},
            clip=false,
            nodes near coords={\pgfmathprintnumber[fixed,precision=2,zerofill]{\pgfplotspointmeta}},
            every node near coord/.append style={font=\fontsize{6}{6.5}\selectfont, anchor=south, yshift=1pt},
            legend image code/.code={\draw[#1, draw=black!65] (0cm,-0.08cm) rectangle (0.18cm,0.08cm);},
            legend style={
                draw=none,
                fill=none,
                font=\scriptsize,
                legend columns=3,
                row sep=1pt,
                column sep=10pt,
                at={(0.5,-0.19)},
                anchor=north
            }
        ]
            \addplot[fill=plotgray, draw=black!65, /pgf/bar shift=0pt] coordinates {(-2.25,21.24) (3.75,20.24) (9.75,19.91)};
            \addplot[fill=rawlight, draw=black!65, /pgf/bar shift=0pt] coordinates {(-1.35,21.51) (4.65,20.24) (10.65,19.91)};
            \addplot[fill=rawdark, draw=black!65, /pgf/bar shift=0pt] coordinates {(-0.45,21.56) (5.55,20.78) (11.55,20.33)};
            \addplot[fill=rgblight, draw=black!65, /pgf/bar shift=0pt] coordinates {(0.45,20.22) (6.45,18.29) (12.45,18.12)};
            \addplot[fill=rgbmid, draw=black!65, /pgf/bar shift=0pt] coordinates {(1.35,20.63) (7.35,17.35) (13.35,18.25)};
            \addplot[fill=rgbdark, draw=black!65, /pgf/bar shift=0pt] coordinates {(2.25,20.92) (8.25,19.39) (14.25,18.61)};
            \legend{Input (no restoration),Strategy A: MOFA (RAWRes),Strategy A: NAFNet (RAWRes),Generic Strategy B: PromptIR,Generic Strategy B: AirNet,Generic Strategy B: MiOIR}
        \end{axis}
    \end{tikzpicture}
    \caption{\textbf{Generic-model comparison, averaged over four device groups and two fixed ISPs.} Strategy A (blue) is trained on clean-noisy RAW images with calibrated degradations, whereas generic Strategy B (orange) uses RGB models pre-trained for all-in-one restoration. This practical transfer comparison is not training-matched.}
    \label{fig:compare}
\end{figure*}
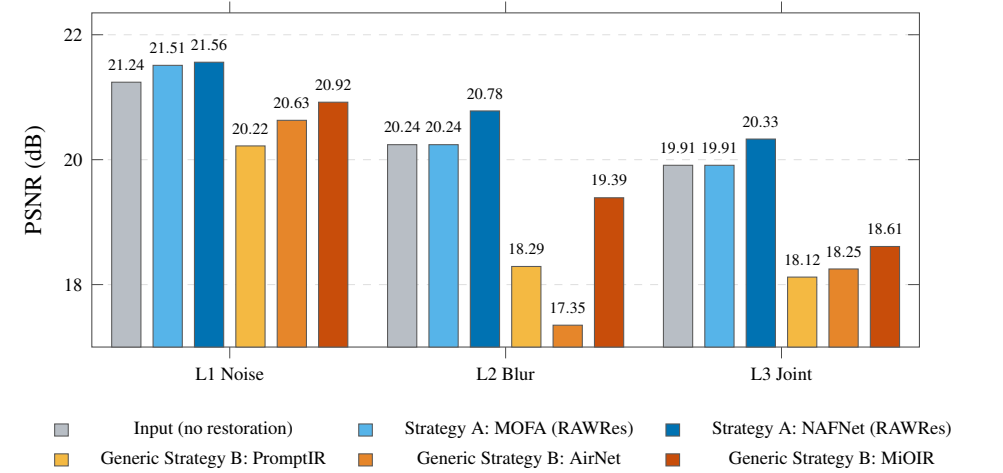

\begin{figure*}[t]
    \centering
    \setlength\tabcolsep{1pt}
    \resizebox{0.98\linewidth}{!}{
    \begin{tabular}{c c c c}
         \includegraphics[width=0.24\linewidth]{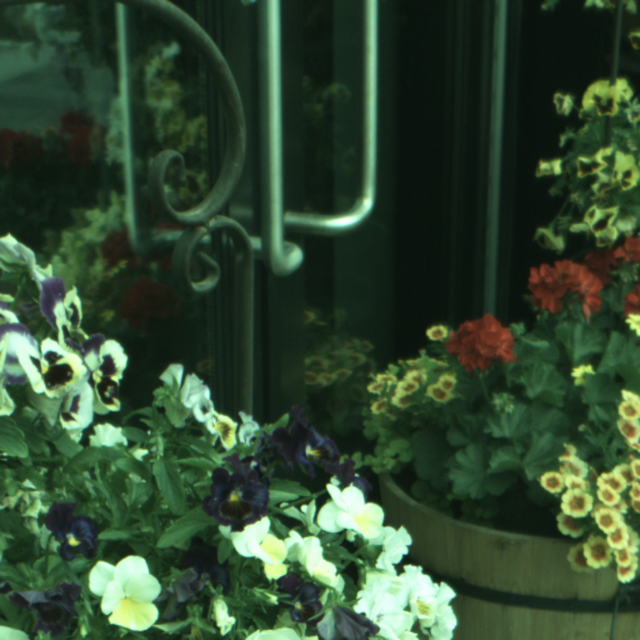} & 
         \includegraphics[width=0.24\linewidth]{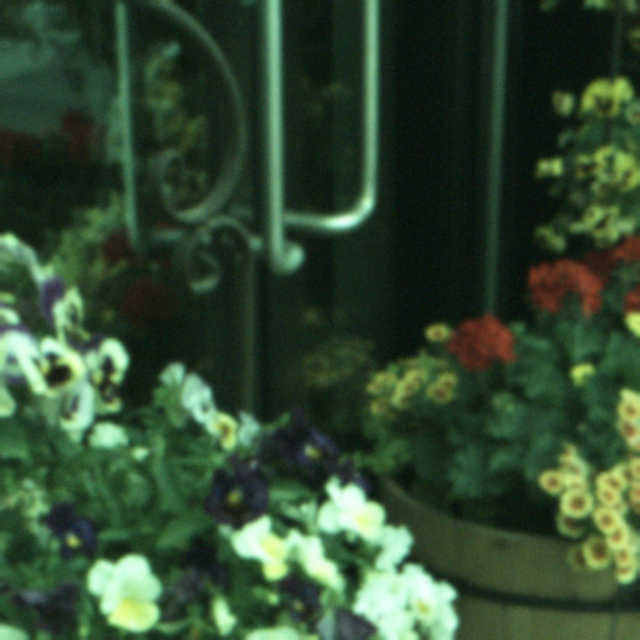} &
         \includegraphics[width=0.24\linewidth]{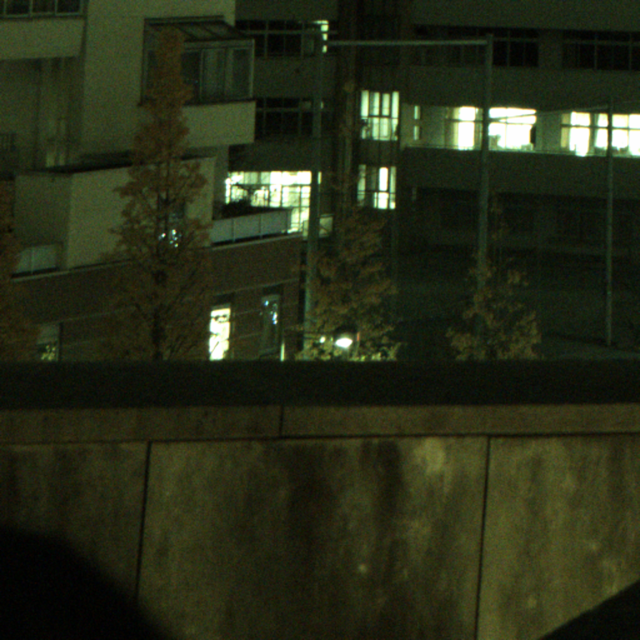} &   
         \includegraphics[width=0.24\linewidth]{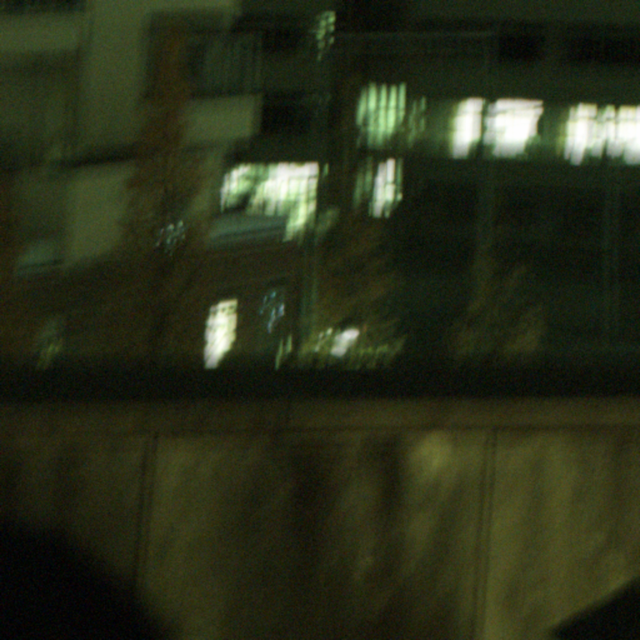} \\
         Clean RAW & Degraded RAW & Clean RAW & Degraded RAW \\
   
         \includegraphics[width=0.24\linewidth]{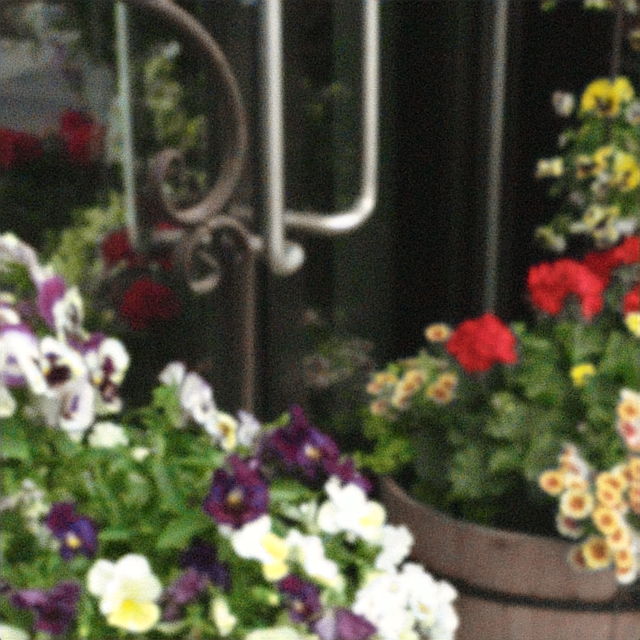} &  
         \includegraphics[width=0.24\linewidth]{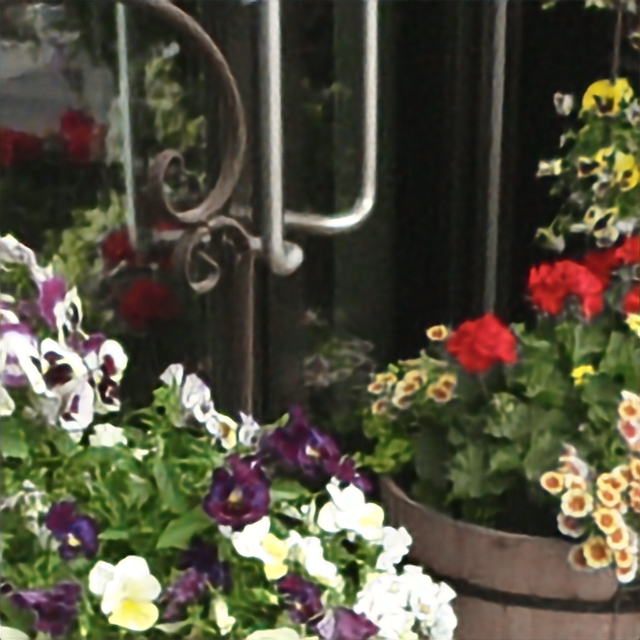} &
         \includegraphics[width=0.24\linewidth]{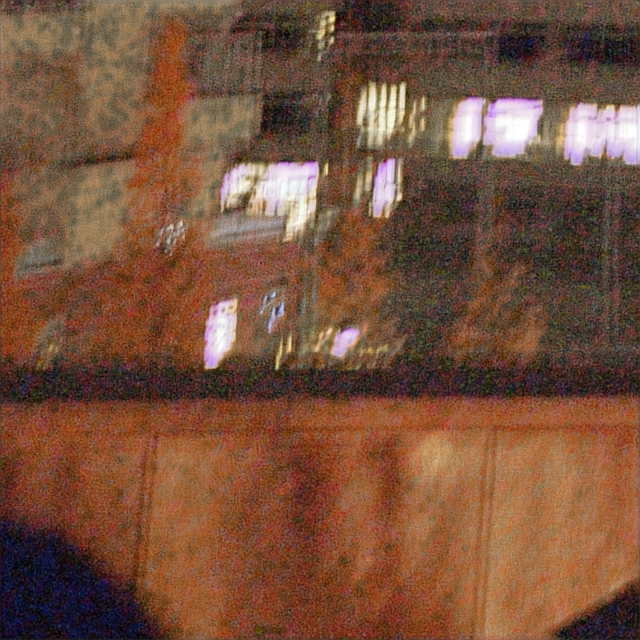} &
         \includegraphics[width=0.24\linewidth]{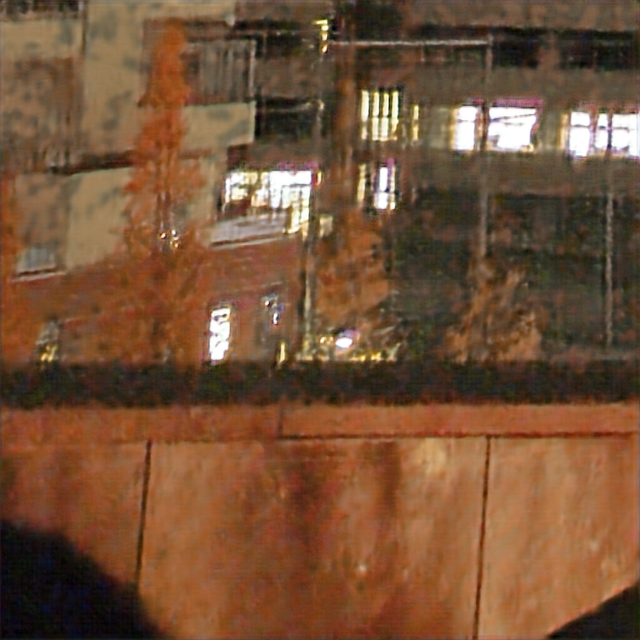} \\

         MiOIR~\cite{kong2024towards} & Ours & MiOIR~\cite{kong2024towards} & Ours \\
         Generic \stb & Generic \sta & Generic \stb & Generic \sta \\
         \\
    \end{tabular}
    }
    \caption{\textbf{RAW--RGB qualitative comparison on the synthetic test set.} We show clean and degraded RAW inputs together with RGB restoration results. The generic MiOIR model~\cite{kong2024towards} receives degraded RGB produced by the fixed ISP, whereas our Strategy A model restores RAW before the same ISP. 
    }
    \label{fig:raw-quality_sample}
\end{figure*}


\subsection{Quantitative Results}
\label{sec:quanti_rlt}

In the quantitative study, we use NAFNet~\cite{chen2022simple} and MOFA~\cite{chen2023mofa} as our \textbf{pre-ISP} models--\emph{RAWRes}-- for \sta. We consider three generic all-in-one RGB restoration models (MiOIR~\cite{kong2024towards}, AirNet~\cite{li2022all}, and PromptIR~\cite{vaishnav2023promptir}) for \stb, together with the two fixed neural ISPs.

The benchmark results in Tables~\ref{tab:robisp} and~\ref{tab:robrgbisp} show that the benchmark-trained RAW models outperform the generic pre-trained RGB models. Figure~\ref{fig:compare} summarizes these tables by averaging PSNR across the four device groups and two fixed ISPs for each degradation level. This comparison demonstrates the limitations of transferring conventional RGB restoration directly to ISP outputs; it does not establish an intrinsic advantage of the RAW domain.

For example, at Level 2 on Samsung S9, \texttt{NAFNet+ISPv1} reaches 25.99 dB, compared with 22.49 dB for \texttt{ISPv1+MiOIR}, a gain of 3.50 dB.
The performance gap against these generic RGB baselines generally widens for the joint degradation at Level 3, indicating that their training distribution does not transfer well to the more complex ISP-transformed artifacts.

Comparing Table~\ref{tab:isp_on_deg_data} with Table~\ref{tab:robisp}, we observe that \emph{pre-ISP} RAW restoration brings the ISP performance closer to the clean baseline. For example, on Vivo X90 at Level 1, the degraded ISP drops from 22.60 dB to 21.52 dB, yet \texttt{NAFNet+ISPv1} recovers to 22.00 dB--only {0.60} dB below the clean performance. This shows that RAW restoration effectively ``shields'' the ISP from input degradations.

Furthermore, the compact RAW models (approximately 1--2 million parameters) obtain these improvements with lower model capacity than the pre-trained RGB baselines (15--30 million parameters), which is favorable for mobile deployment.

\paragraph{Target-ISP Training} Figure~\ref{fig:x90compare} shows a different regime of \stb: the post-ISP model is trained on the target ISP output distribution. With comparable model size, this model exceeds the sensor-specific \sta result across all three degradation levels. It can learn artifacts present in the final RGB output while the ISP remains fixed and is treated as a black box. Together with the generic-model comparison, this reversal supports our main conclusion that training-distribution alignment, rather than domain alone, determines the observed ranking.

\subsection{Qualitative Results}
\label{sec:quali_rlt}


In the qualitative studies, we use NAFNet~\cite{chen2022simple} as the backbone of the \textbf{RAWRes} pre-ISP block because it is the strongest RAW baseline in Table~\ref{tab:model_baseline}.

\begin{figure*}[!htp]
    \centering
    \setlength\tabcolsep{2pt}
    \resizebox{0.98\linewidth}{!}{
        \begin{tabular}{c c c c}
    
             \includegraphics[width=0.24\linewidth]{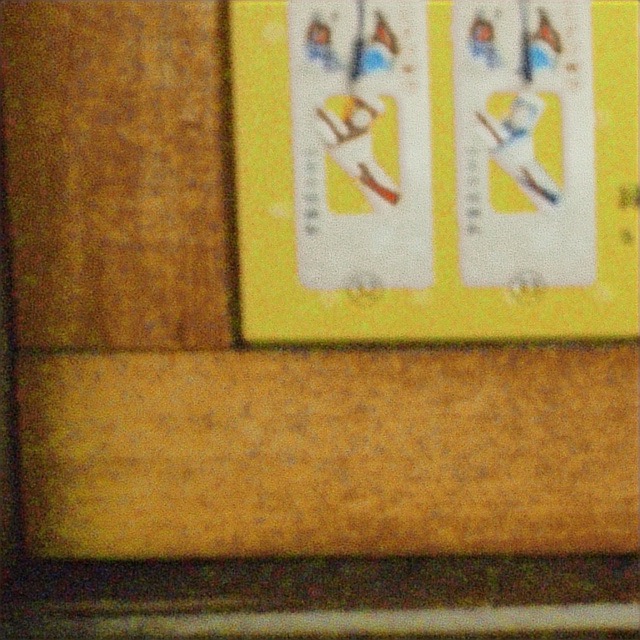} &
             \includegraphics[width=0.24\linewidth]{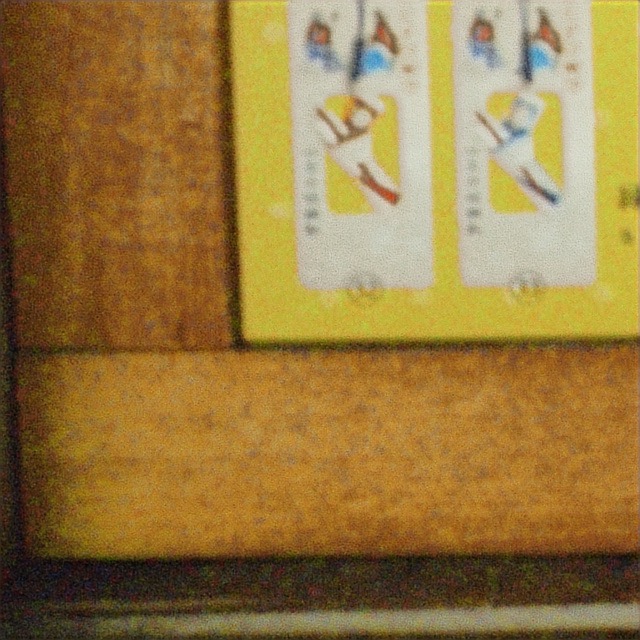} & 
             \includegraphics[width=0.24\linewidth]{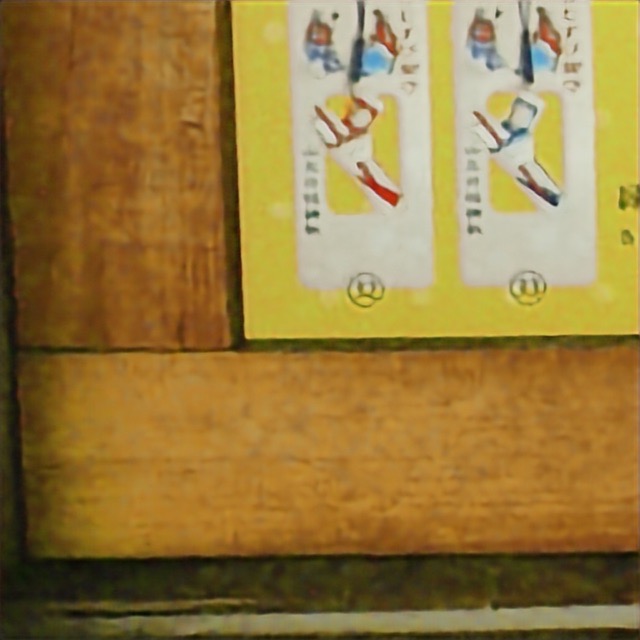} & 
             \includegraphics[width=0.24\linewidth]{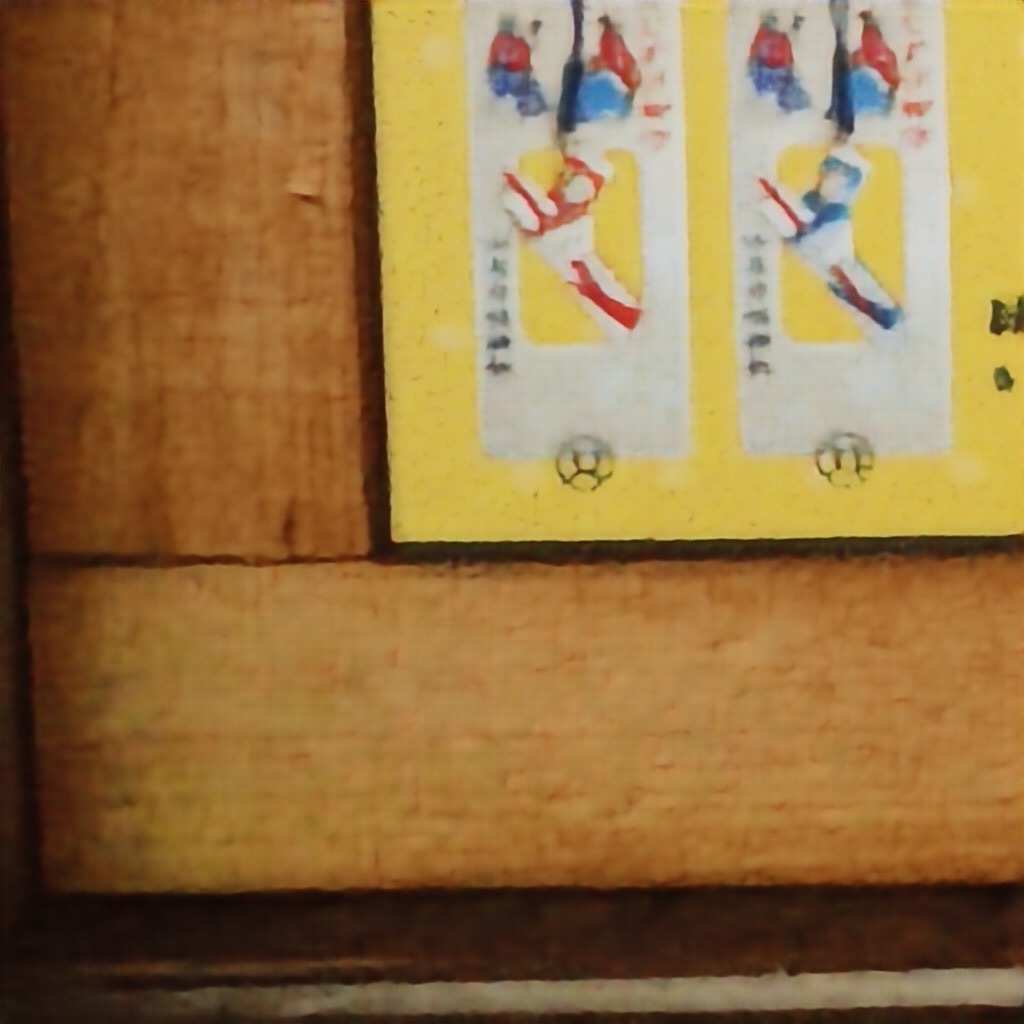} \\ 
                 
             \includegraphics[width=0.24\linewidth]{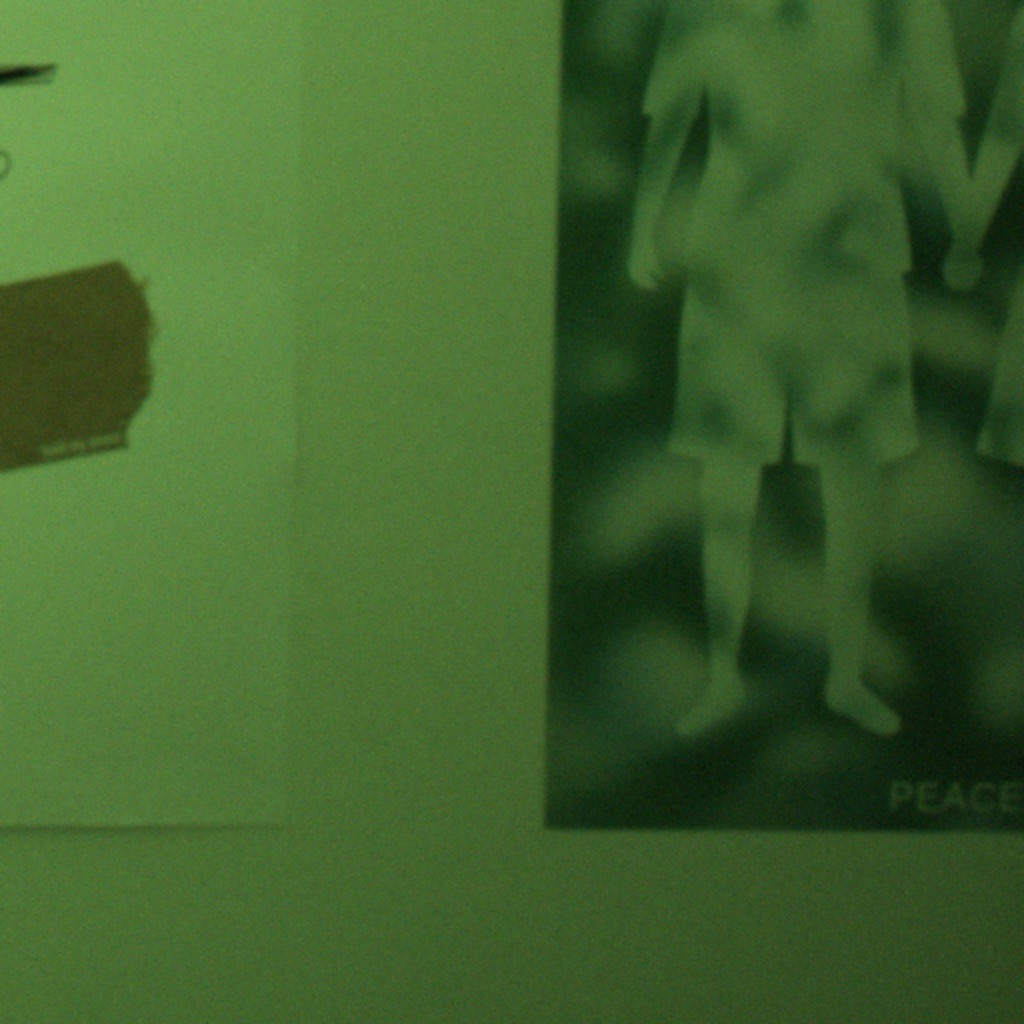} &
             \includegraphics[width=0.24\linewidth]{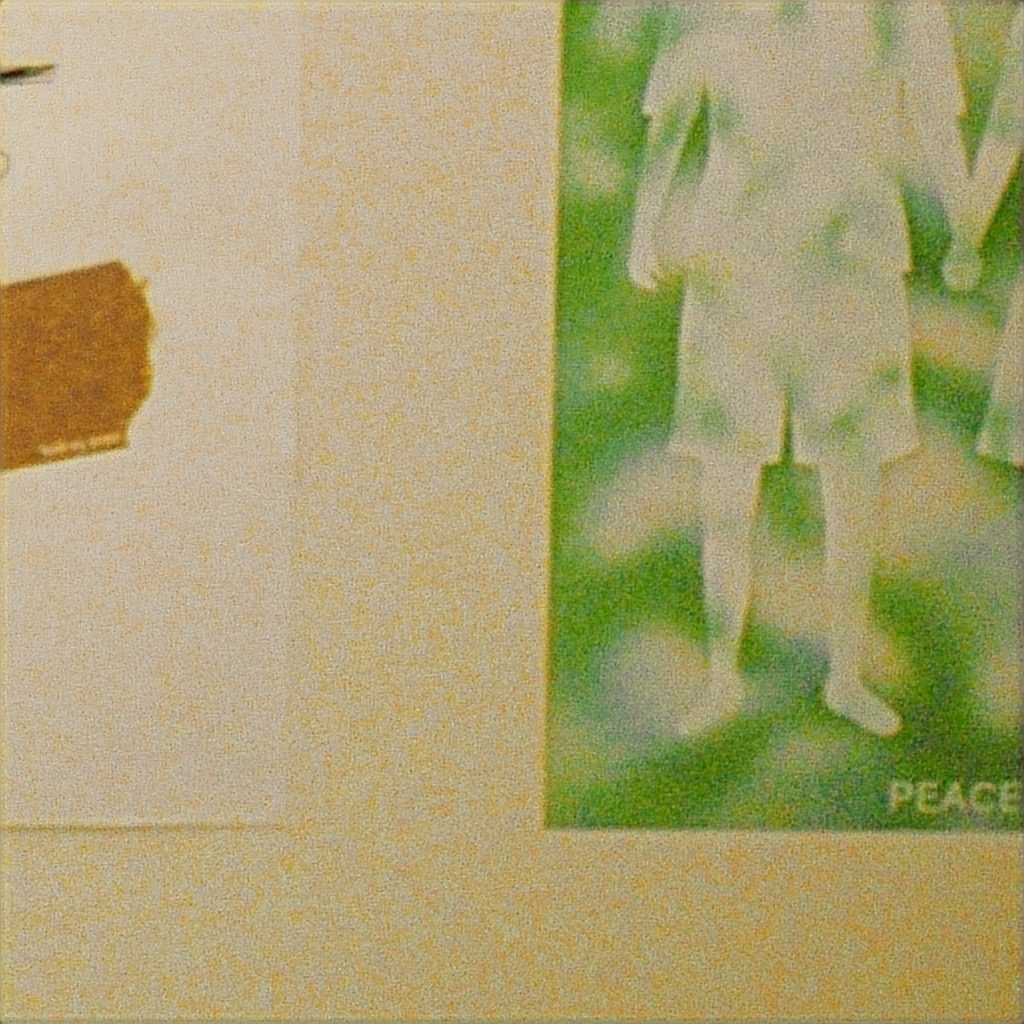} & 
             \includegraphics[width=0.24\linewidth]{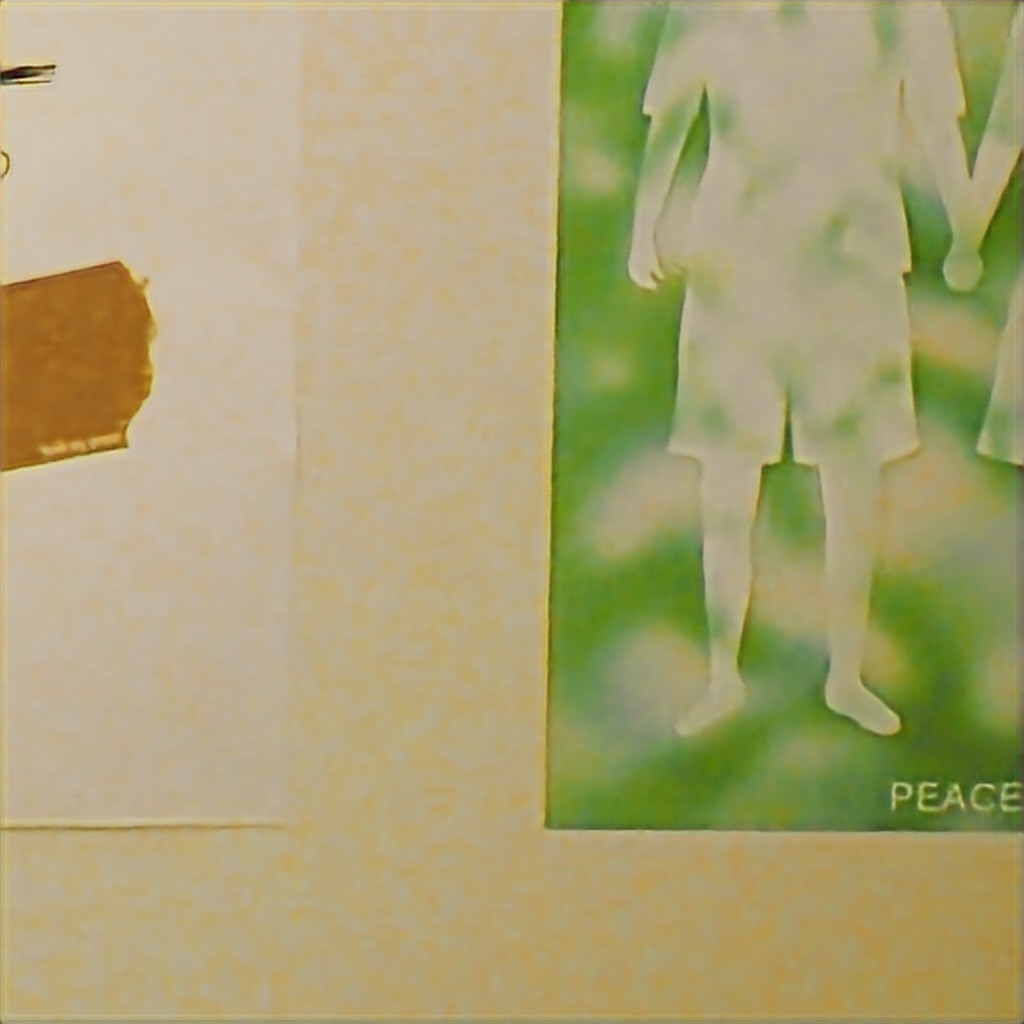} & 
             \includegraphics[width=0.24\linewidth]{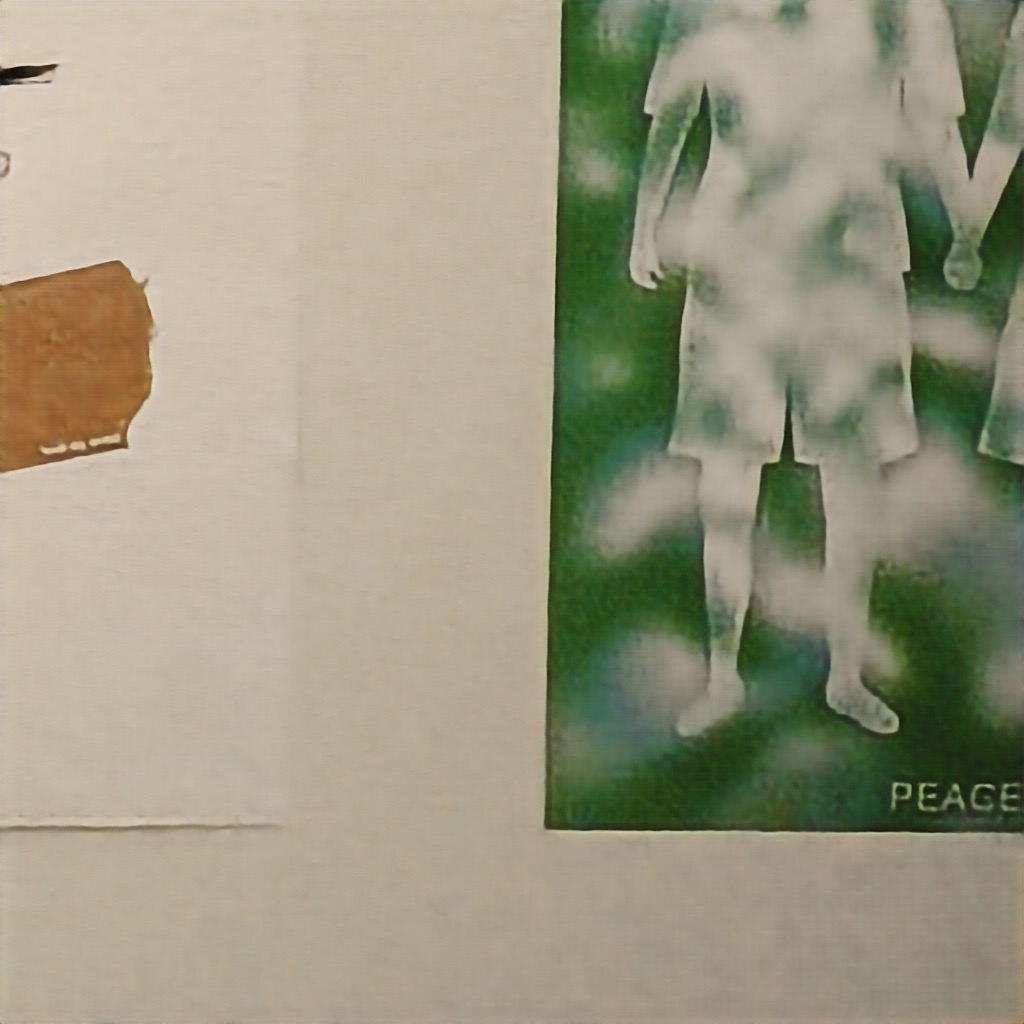} \\ 
             
             \includegraphics[width=0.24\linewidth]{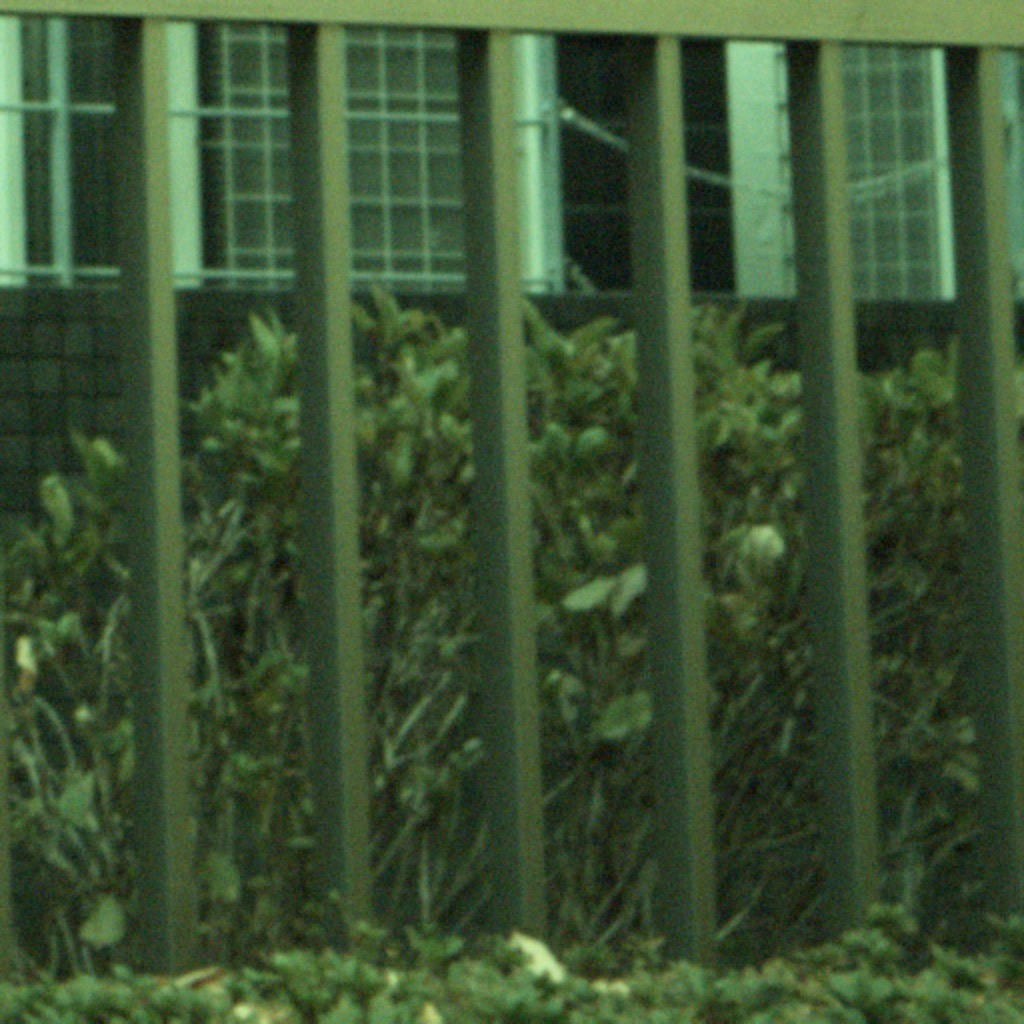} &
             \includegraphics[width=0.24\linewidth]{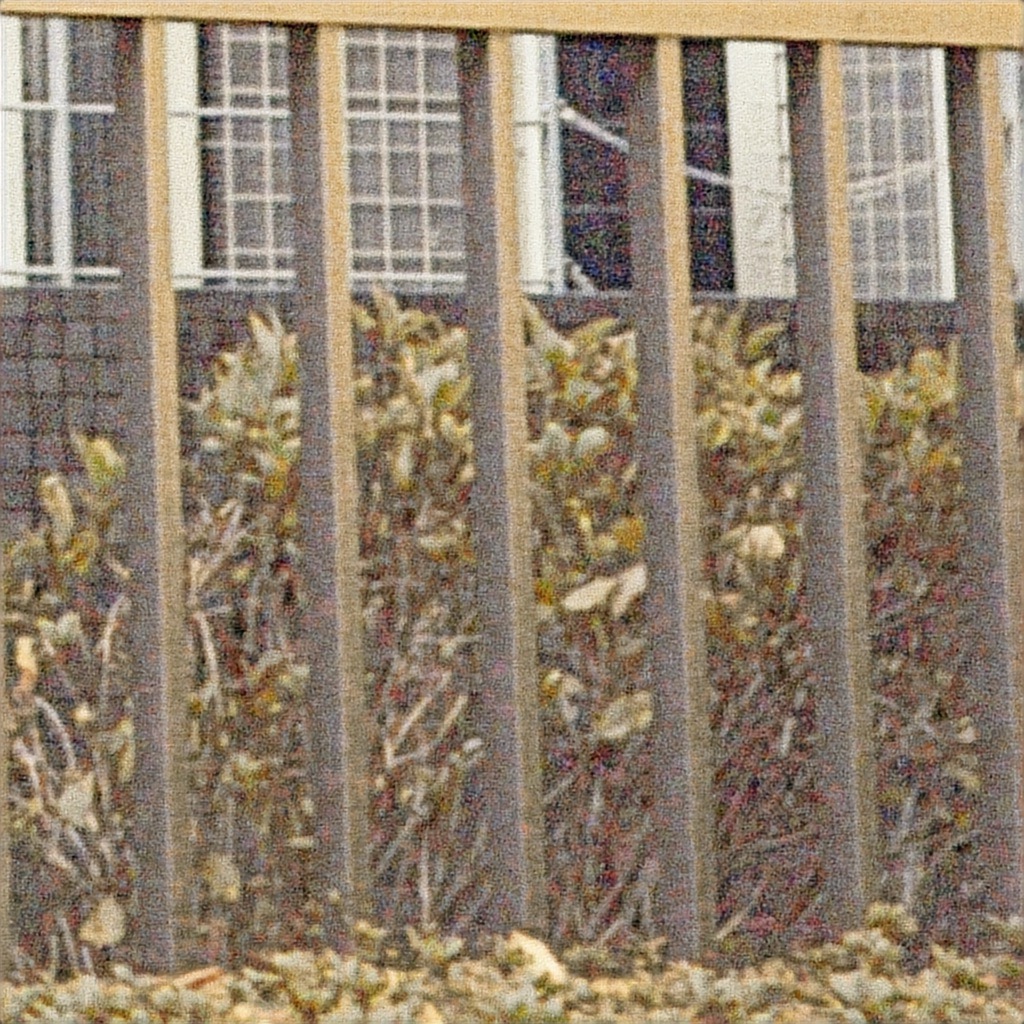} & 
             \includegraphics[width=0.24\linewidth]{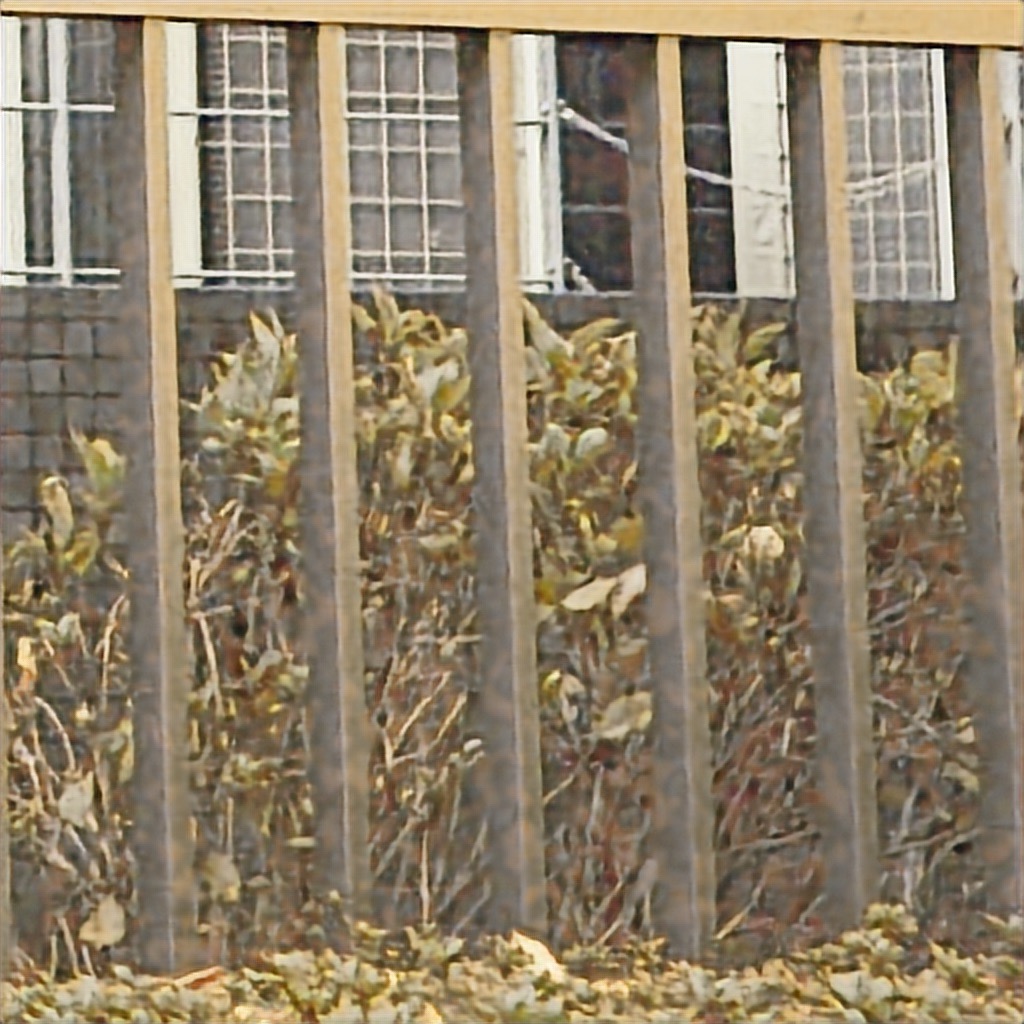} & 
             \includegraphics[width=0.24\linewidth]{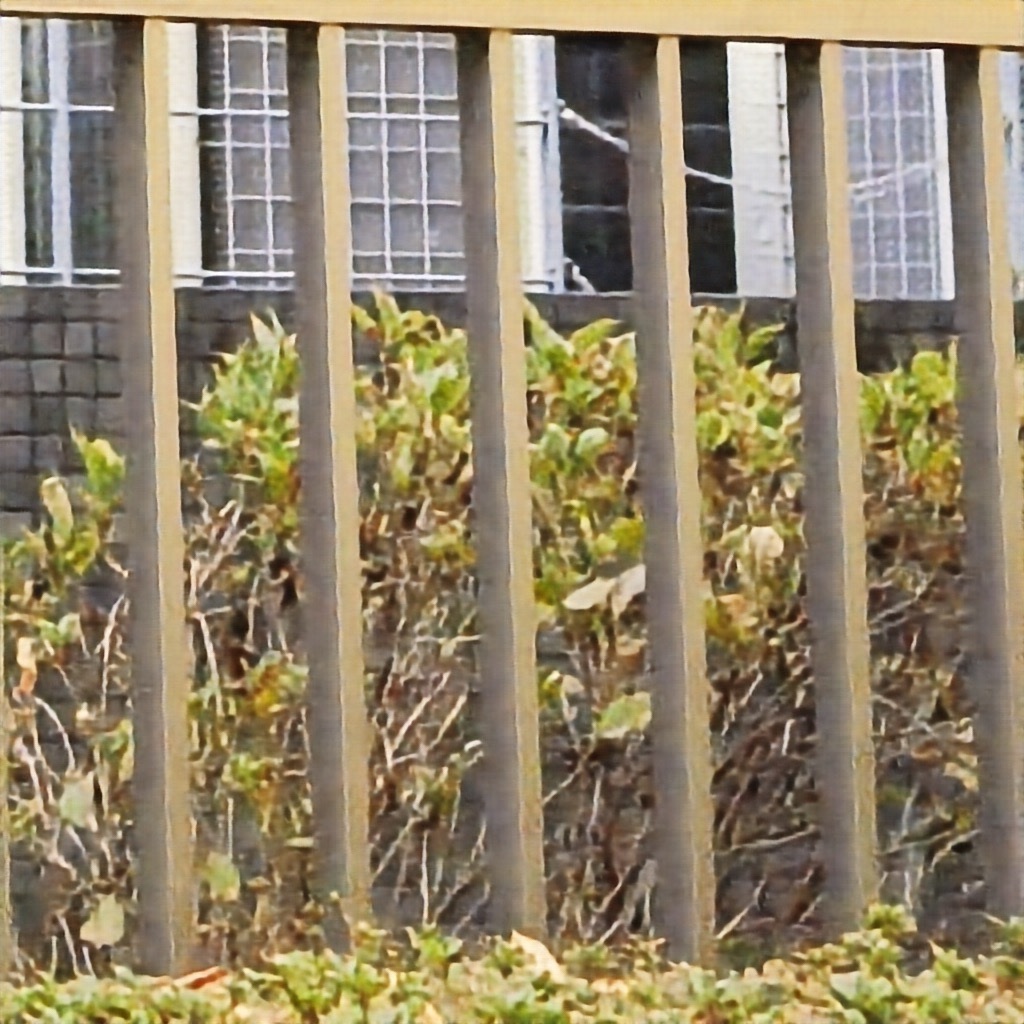} \\ 
             
             \includegraphics[width=0.24\linewidth]{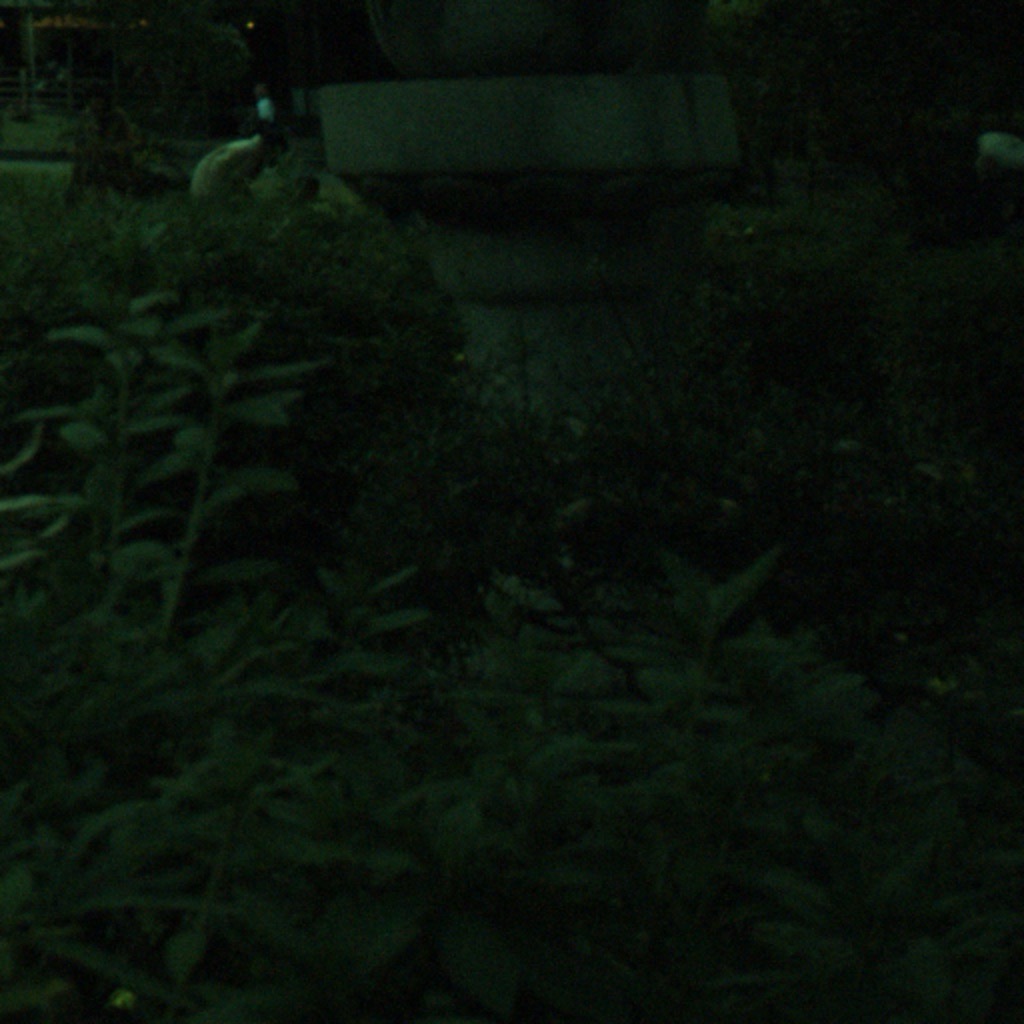} &
             \includegraphics[width=0.24\linewidth]{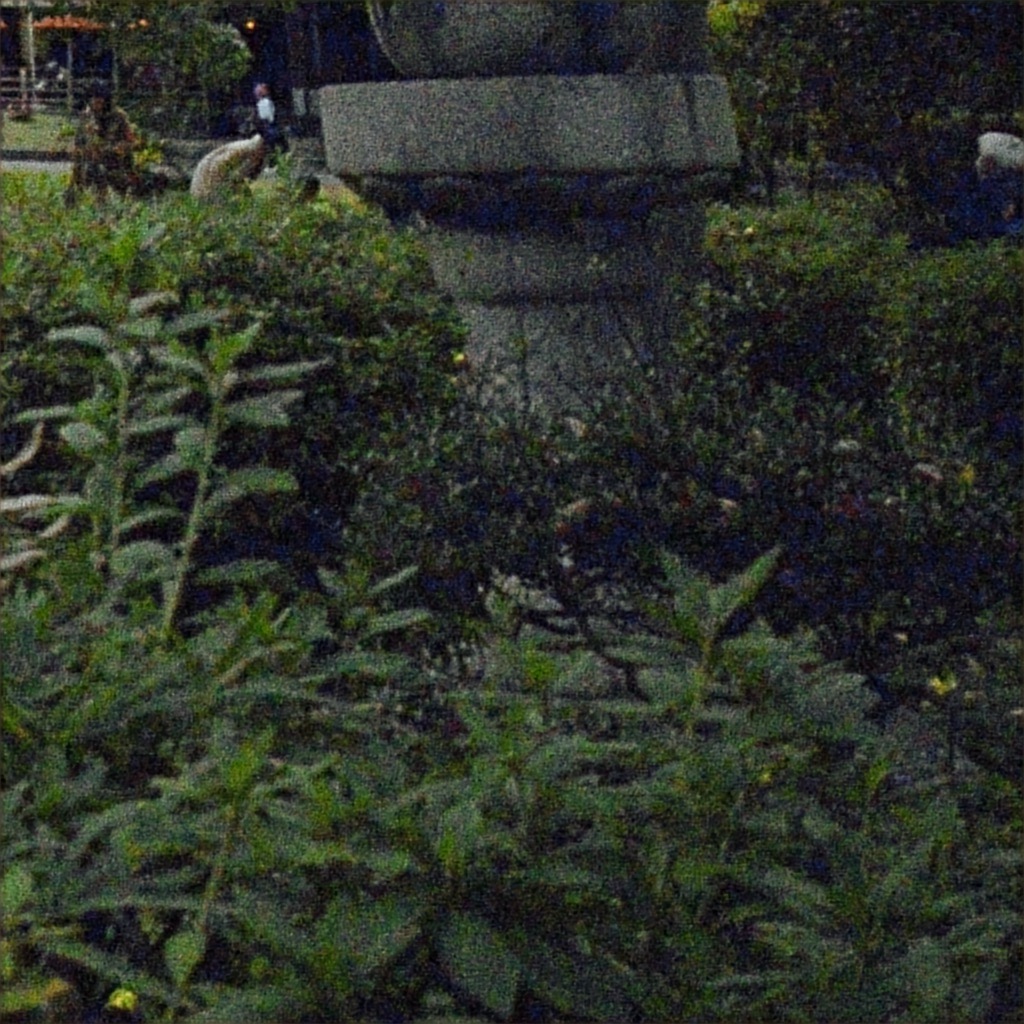} & 
             \includegraphics[width=0.24\linewidth]{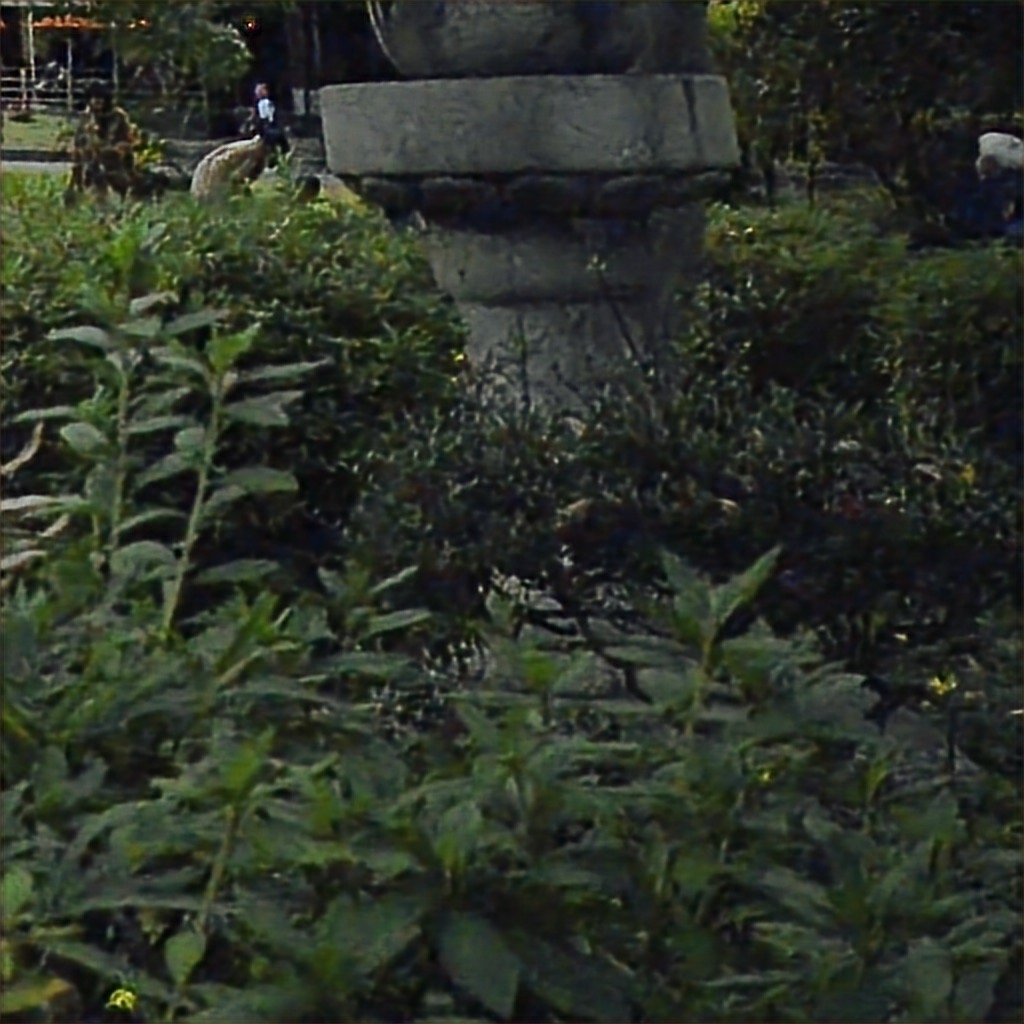} & 
             \includegraphics[width=0.24\linewidth]{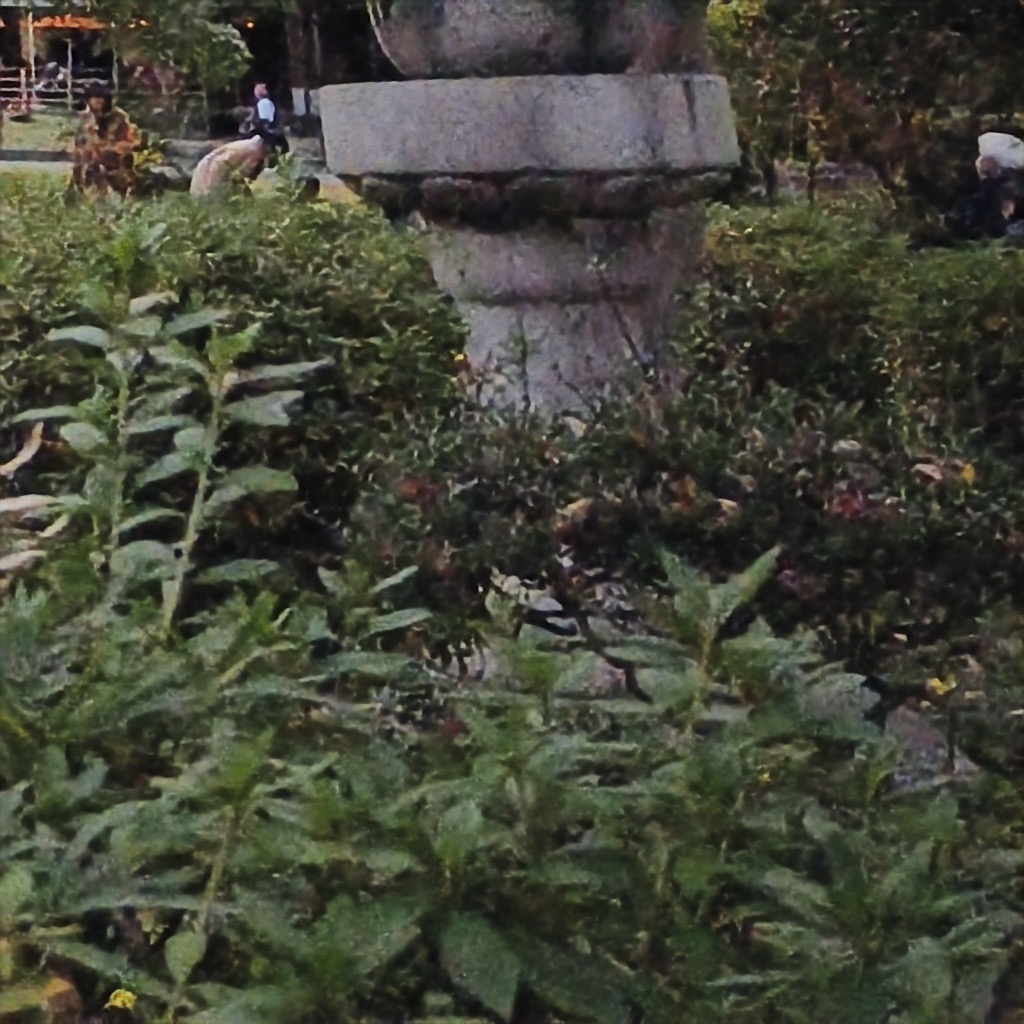} \\ 
             
             \includegraphics[width=0.24\linewidth]{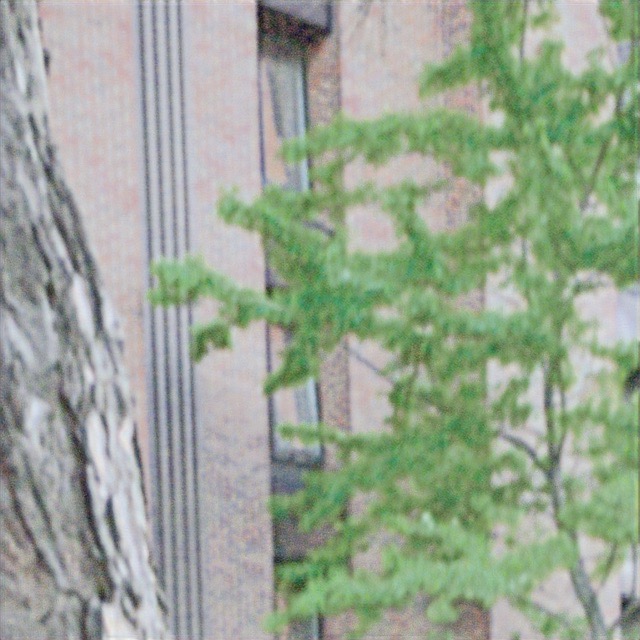} &
             \includegraphics[width=0.24\linewidth]{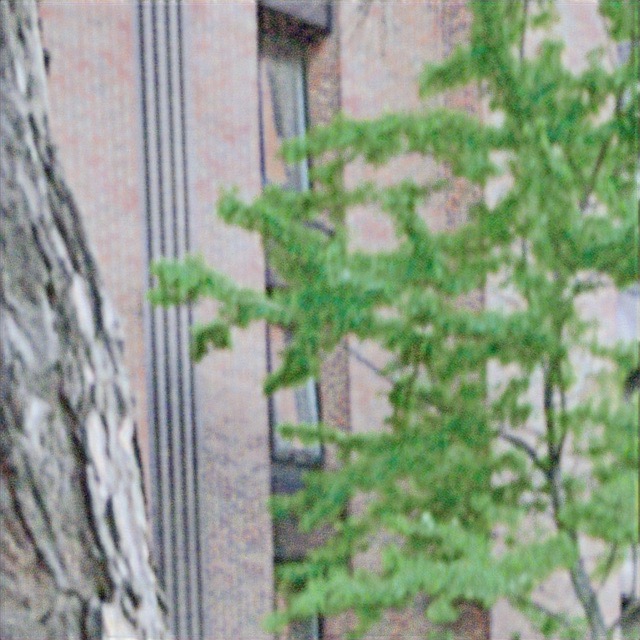} & 
             \includegraphics[width=0.24\linewidth]{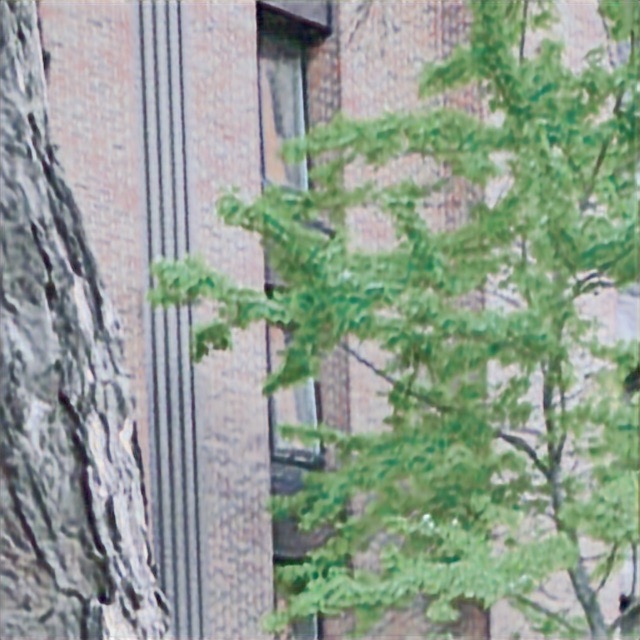} &
             \includegraphics[width=0.24\linewidth]{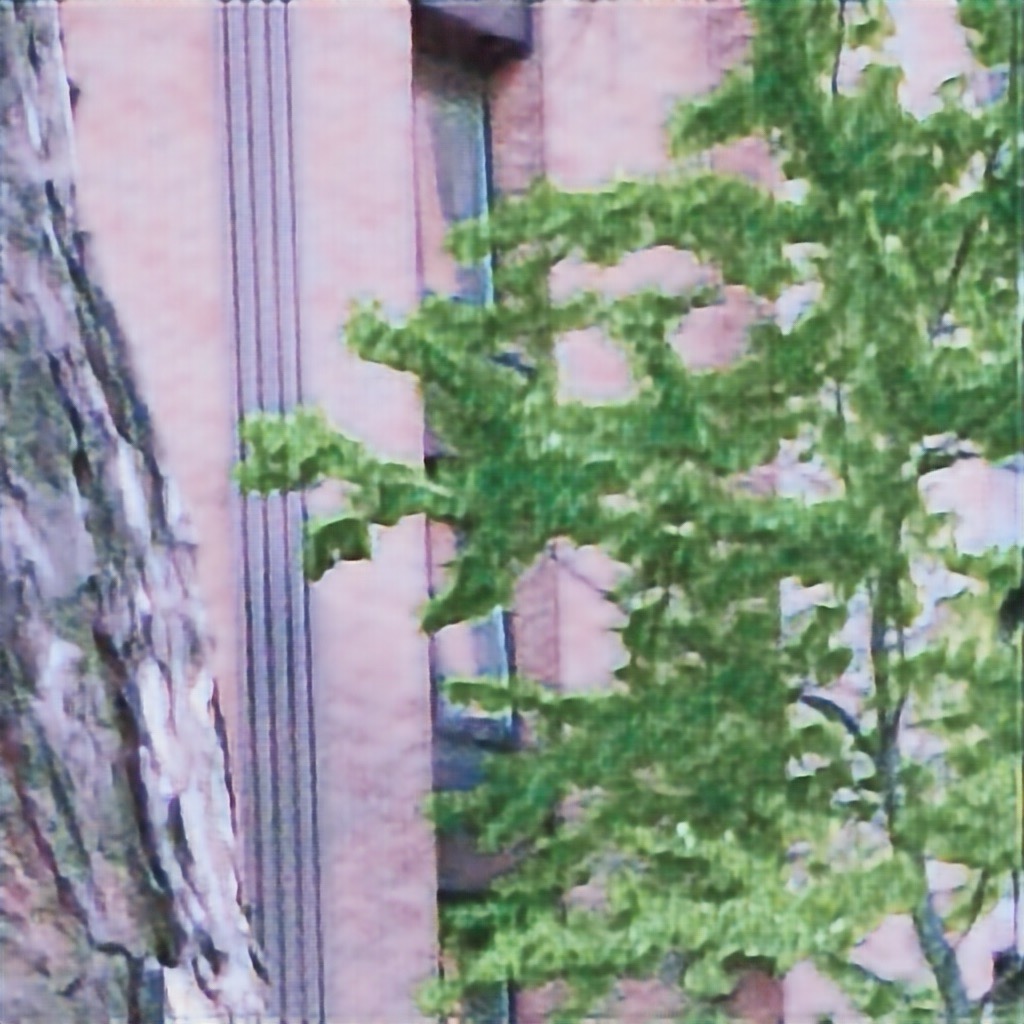} \\
             Degraded RGB & MiOIR~\cite{kong2024towards}  & Ours   & Ours  \\
             & Generic \stb & Generic \sta & ISP-Aware \stb \\
             \\
        \end{tabular}
    }
    \caption{\textbf{Qualitative comparison on the synthetic test set.} MiOIR~\cite{kong2024towards} receives RGB images produced by passing degraded RAW data through the fixed ISP. Both our pre-ISP RAW model and our target-ISP-trained post-ISP RGB model reduce visible blur and noise. In these examples, target-ISP-trained RGB restoration recovers more color and detail.
    }
    \label{fig:rgb_quality_sample5}
\end{figure*}

Figures~\ref{fig:raw-quality_sample} and~\ref{fig:rgb_quality_sample5} compare our trained models with the generic RGB restoration baseline MiOIR~\cite{kong2024towards}, which is generally the strongest pre-trained RGB method in Table~\ref{tab:robrgbisp}. We use the same fixed ISP~v1 for these comparisons.

The visual examples show that both our pre-ISP and post-ISP-aware models reduce noise and motion blur more effectively than MiOIR~\cite{kong2024towards} on the selected samples.

Moreover, as discussed in Sec.~\ref{sec:quanti_rlt}, the \emph{target-ISP-trained post-ISP}  model can restore artifacts present after ISP processing. This result is specific to training on the target output distribution and should not be generalized to arbitrary RGB restoration models, as illustrated by the weaker generic baselines.

\begin{figure*}[!ht]
    \centering
    \setlength\tabcolsep{2pt}
    \resizebox{\linewidth}{!}{
        \begin{tabular}{c c c}
             \includegraphics[width=0.32\linewidth]{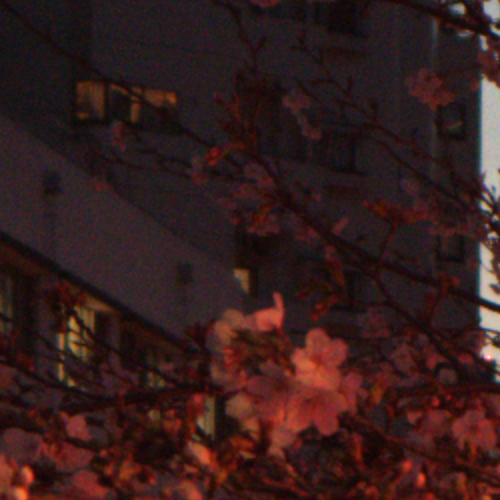} & 
             \includegraphics[width=0.32\linewidth]{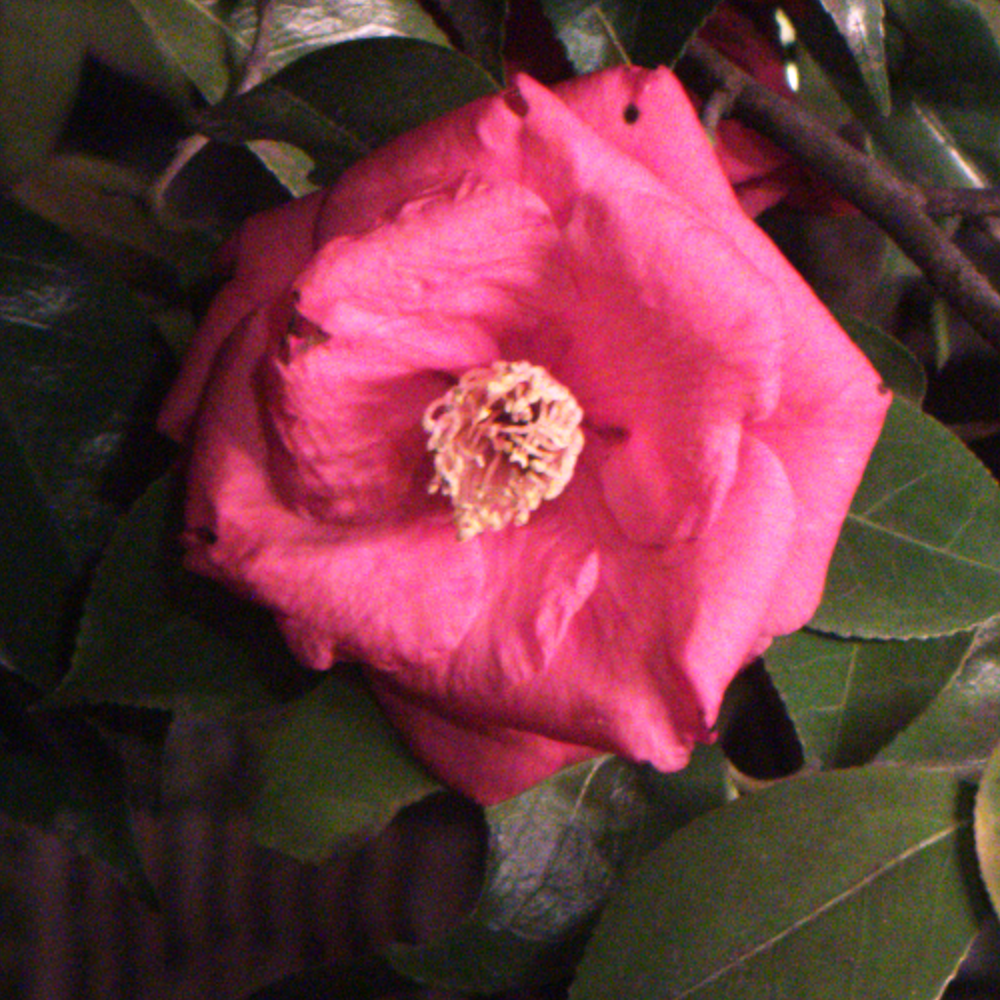} & 
             \includegraphics[width=0.32\linewidth]{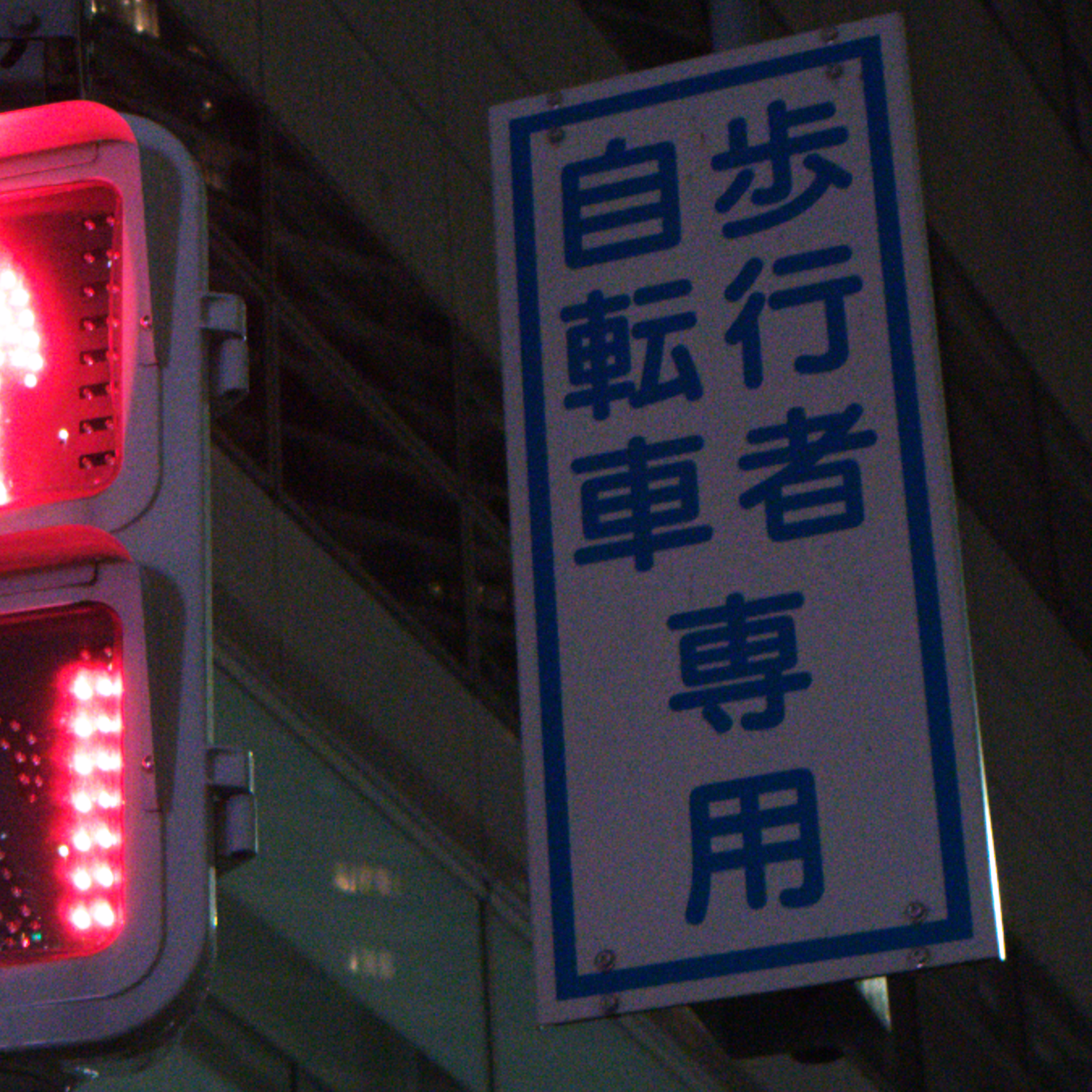} \\

             \multicolumn{3}{c}{Original Vivo X90 camera output} \\
             \includegraphics[width=0.32\linewidth]{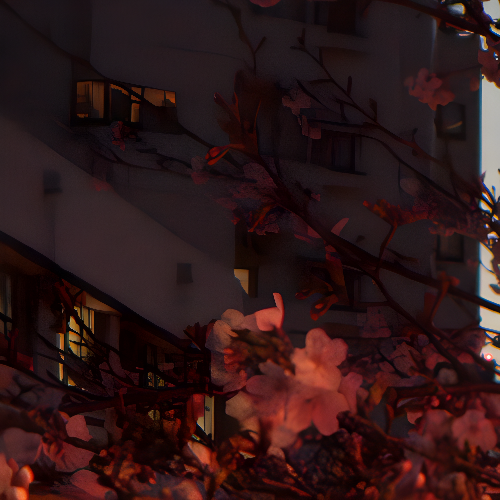} & 
             \includegraphics[width=0.32\linewidth]{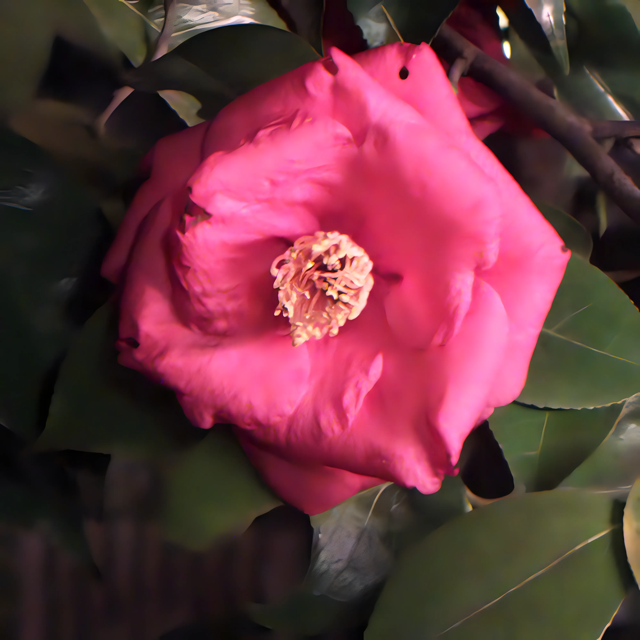} & 
             \includegraphics[width=0.32\linewidth]{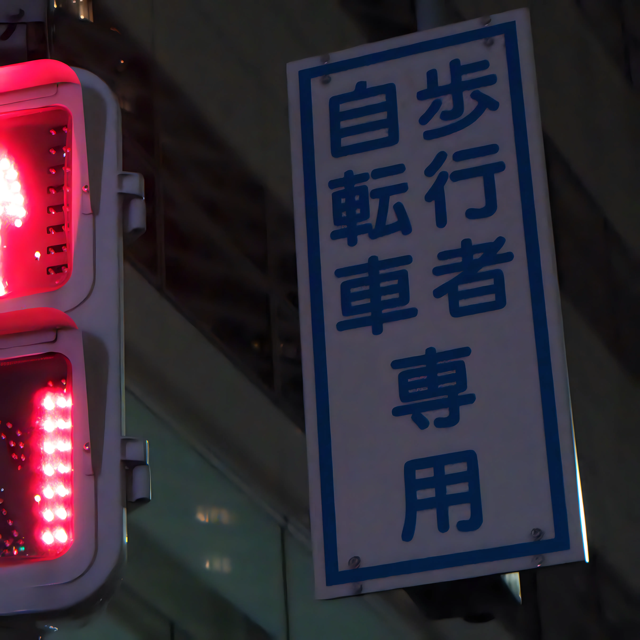} \\

             \multicolumn{3}{c}{RealESRGAN results after processing (A)} \\
    
             \includegraphics[width=0.32\linewidth]{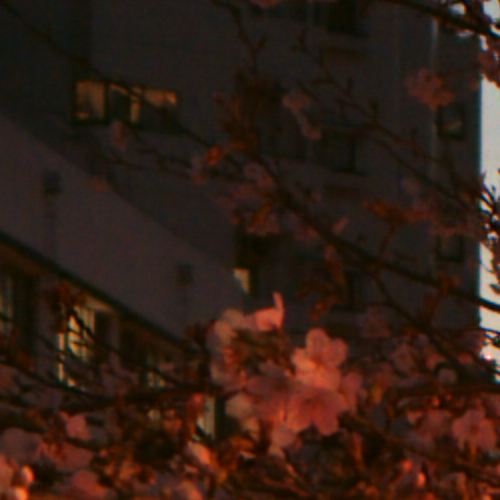} & 
             \includegraphics[width=0.32\linewidth]{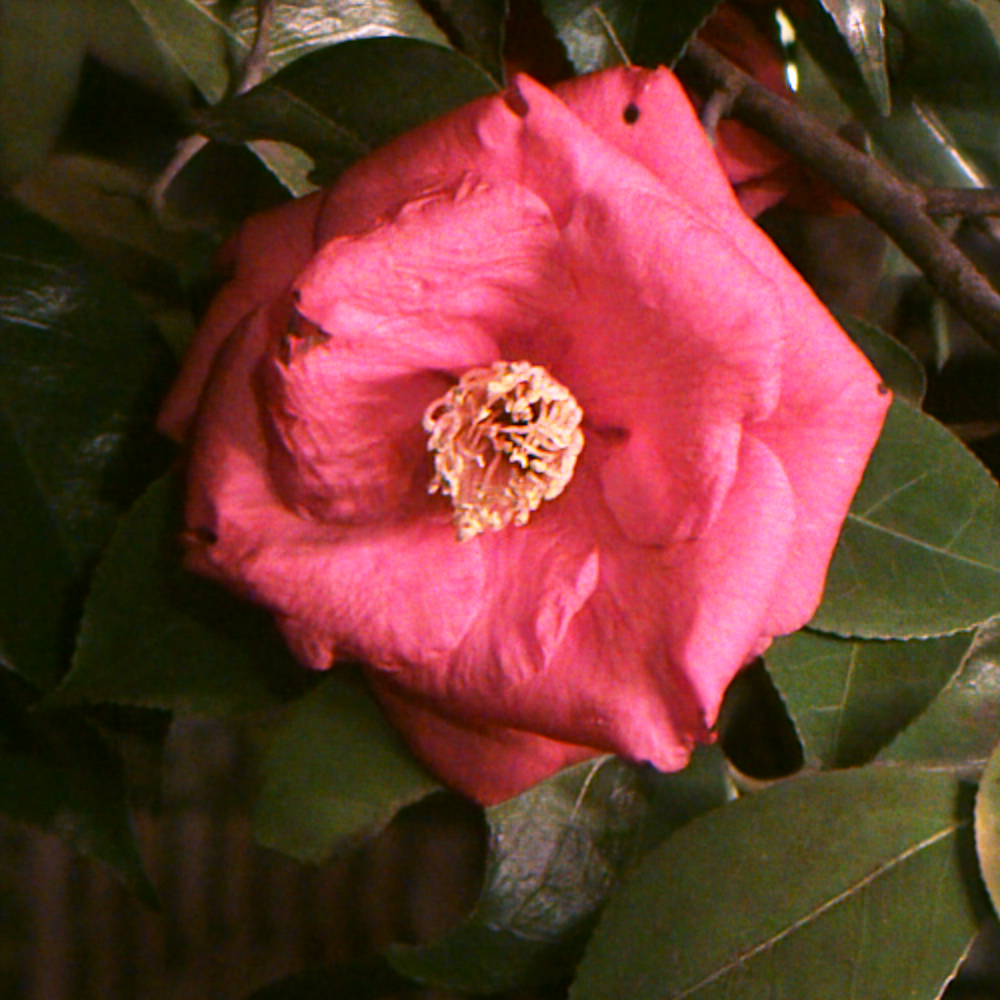} & 
             \includegraphics[width=0.32\linewidth]{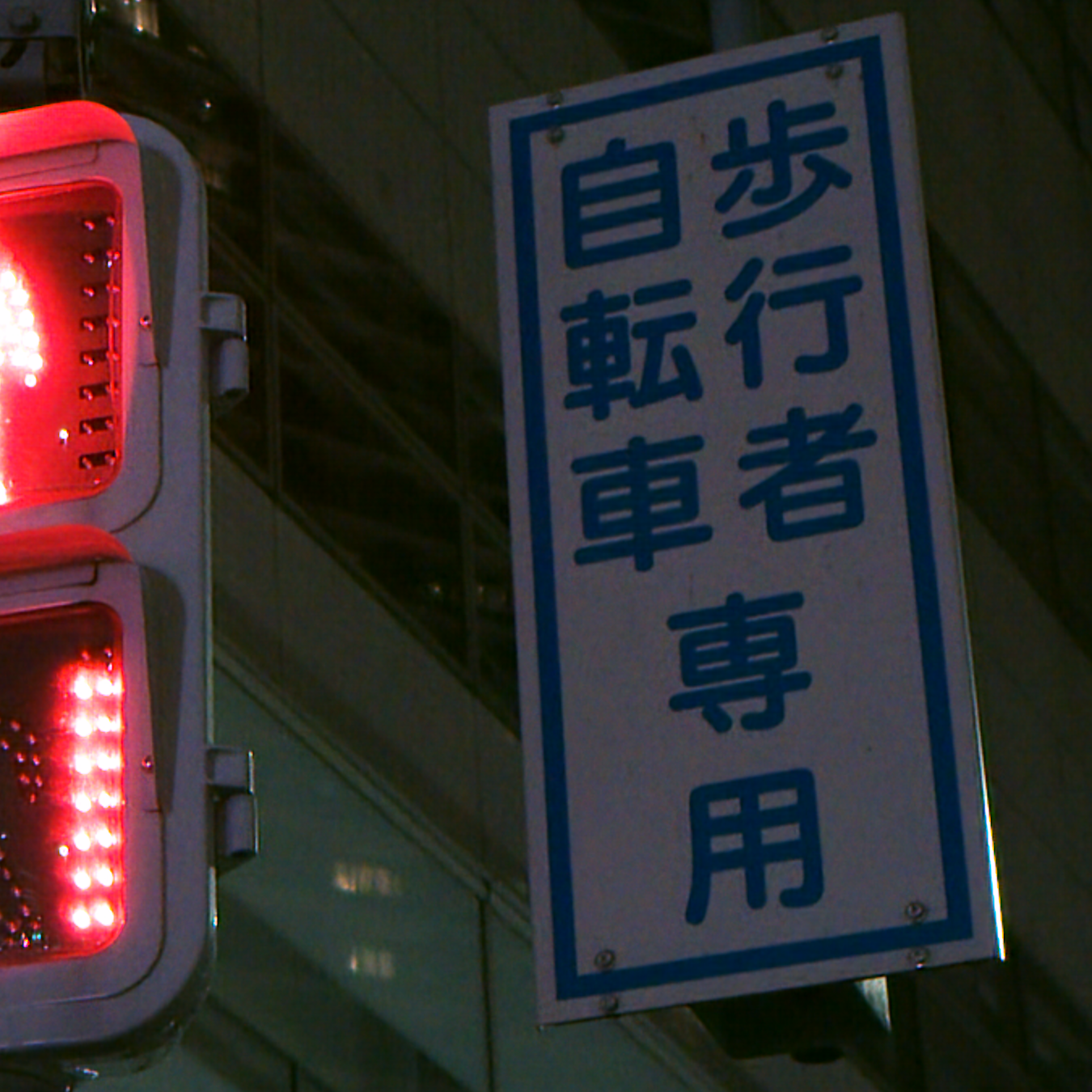} \\

             \multicolumn{3}{c}{Our \stb: ISP-aware RGB restoration method}
             \\
             \\
        \end{tabular}
    }
    \caption{\textbf{Real-world qualitative results.} We use the same RAW images and camera pipeline for all comparisons. Our target-ISP-trained RGB restoration reduces visible noise while preserving fine details in these examples.}
    \label{fig:realworld}
\end{figure*}

\vspace{-2mm}
\paragraph{Real-World Evaluation} 
We apply our pipeline to real-world RAW images captured using a Vivo X90 phone in challenging conditions. We obtain the sRGB images directly from the smartphone's ISP. In Figure~\ref{fig:realworld}, the original phone output exhibits strong sensor noise and blur. Our ISP-aware RGB restoration model produces images with less visible noise and clearer details. We also compare with RealESRGAN~\cite{wang2021realESRGAN}, which produces oversmoothed textures in these examples. This illustrates a potential risk of deploying generative restoration models in photography pipelines.

\subsection{Limitations and Future Work}
\label{sec:limit}

First, although the RAW degradation pipeline aims to approximate real-world degradations, synthetic noise and blur cannot capture every property of smartphone sensors and lenses. Together with the 47-image test split, this limits broad claims about generalization.

\vspace{2mm}

\noindent Second, the generic comparison is not training-matched, and the sensor-specific RAW and RGB models optimize losses in different domains. The reported reversal therefore supports the importance of target-distribution alignment, but it does not determine the intrinsic performance ceiling of either domain.

\vspace{2mm}

\noindent Third, we do not investigate Color Correction Matrix-aware or target-ISP-aware RAW objectives. Injecting CCM priors or optimizing RAW restoration through a suitable approximation of the downstream pipeline could improve Strategy A and change the observed ranking. We consider this an important direction for future work rather than a conclusion supported by the current benchmark.

\vspace{2mm}

\noindent Finally, larger real-world test sets and additional modern cameras are needed to assess generalization beyond the evaluated sensors and pipeline proxies.

\section{Conclusion}
\label{sec:conclusion}


We benchmark deep restoration before and after a fixed image signal processor using four smartphone device groups, two learned ISP proxies, and representative RAW and RGB restoration methods. Benchmark-trained RAW restoration outperforms generic pre-trained RGB restoration in most evaluated configurations, whereas target-ISP-trained RGB restoration performs best in the Vivo X90 pipeline-specific experiment. These results do not identify a universally preferable restoration domain. Instead, they show that matching the restoration training distribution to the distribution induced at the model's insertion point in the target imaging pipeline is central to robust final-image quality.

\subsection*{Acknowledgments} 
This work was partly supported by the Alexander von Humboldt Foundation.


\bibliography{egbib}
\end{document}